\documentclass[journal,comsoc]{IEEEtran}

\usepackage[T1]{fontenc}

\usepackage{graphicx}
\usepackage{amsmath}
\usepackage{amssymb}
\usepackage{color}
\usepackage[table]{xcolor}
\usepackage[misc]{ifsym}
\usepackage{subfigure}
\usepackage{array}
\usepackage{enumerate}
\usepackage{multirow}
\usepackage{verbatim}
\usepackage{cite}
\usepackage{url}
\usepackage{algorithm}
\usepackage{algorithmic}
\usepackage{amsmath}
\usepackage{verbatim}
\usepackage{color}
\ifCLASSINFOpdf

\else

\fi

\usepackage{amsmath}

\usepackage[cmintegrals]{newtxmath}
\usepackage{algorithm}
\usepackage{algorithmic}
\usepackage{amsmath}
\usepackage{verbatim}

\begin{document}

\title{A comprehensive review of remote sensing in wetland classification and mapping}

\author{Shuai Yuan, Xiangan Liang, Tianwu Lin, Shuang Chen, Rui Liu, \\ Jie Wang, Hongsheng Zhang, Peng Gong
\thanks{Shuai Yuan, Shuang Chen, Rui Liu, Hongsheng Zhang, Peng Gong are with the Department of Geography, The University of Hong Kong, Hong Kong, China (Email: shuai914@connect.hku.hk, schen17@connect.hku.hk, rhysliu@connect.hku.hk, zhanghs@hku.hk, penggong@hku.hk).}
\thanks{Xiangan Liang is with the Department of Earth System Science, Tsinghua University, Beijing, China (Email: lxa21@mails.tsinghua.edu.cn).}
\thanks{Tianwu Lin is with the Department of Electronics and Information Engineering, Harbin Institute of Technology (Shenzhen), Shenzhen, China and Pengcheng Laboratory, Shenzhen, China (Email: lintw@stu.hit.edu.cn).}
\thanks{Jie Wang is with Pengcheng Laboratory, Shenzhen, China (Email: wangj10@pcl.ac.cn)}
\thanks{\textit{Corresponding authors: Peng Gong and Jie Wang.}}
        }

\maketitle

\begin{abstract}
Wetlands constitute critical ecosystems that support both biodiversity and human well-being; however, they have experienced a significant decline since the 20th century. Back in the 1970s, researchers began to employ remote sensing technologies for wetland classification and mapping to elucidate the extent and variations of wetlands. Although some review articles summarized the development of this field, there is a lack of a thorough and in-depth understanding of wetland classification and mapping: (1) the scientific importance of wetlands, (2) major data, methods used in wetland classification and mapping, (3) driving factors of wetland changes, (4) current research paradigm and limitations, (5) challenges and opportunities in wetland classification and mapping under the context of technological innovation and global environmental change. 
In this review, we aim to provide a comprehensive perspective and new insights into wetland classification and mapping for readers to answer these questions. First, we conduct a meta-analysis of over 1,200 papers, encompassing wetland types, methods, sensor types, and study sites, examining prevailing trends in wetland classification and mapping. Next, we review and synthesize the wetland features and existing data and methods in wetland classification and mapping. We also summarize typical wetland mapping products and explore the intrinsic driving factors of wetland changes across multiple spatial and temporal scales. Finally, we discuss current limitations and propose future directions in response to global environmental change and technological innovation. This review consolidates our understanding of wetland remote sensing and offers scientific recommendations that foster transformative progress in wetland science.

\end{abstract}


\IEEEpeerreviewmaketitle

\section{Introduction}
\label{sec:intro}
As unique and crucial ecosystems on Earth, wetlands play an irreplaceable role in maintaining ecological balance, providing ecological services, protecting biodiversity, and promoting human well-being \cite{gong2010china, dronova2015mapping}. Wetlands also exhibit excellent functions in regulating climate, flood control and drought relief, methane emission estimation, etc \cite{zhang2021development, erwin2009wetlands, pattison2018wetlands}. Nevertheless, due to the sensitivity of wetlands to human climate change, over 50\% of wetland areas have either disappeared or degraded since the 20th century, placing them in significant jeopardy.

Wetland classification and mapping provide significant data support and prior understanding for scientific, ecological protection of wetland resources by accurately and timely depicting the types, extents, distribution and dynamic changes of wetlands. Compared with other wetland survey methods that incur high labor costs in a non-timely manner with destruction to wetlands, remote sensing can macroscopically and rapidly acquire data of the wetlands' surface in different bands, observing the spectral, spatial, scattering and even phenological characteristics of wetlands clearly and dynamically in a long time-series period and at a large scale. With the beginning of the first wetland remote sensing research in the 1970s \cite{cowardin1974remote, brown1978wetland, klemas1977remote}, and the development of remote sensing technology since, now there are over 1,200 wetland classification and mapping-related research publications, involving with multi-source data observation and integration, advanced computer technologies, and providing important scientific support for various downstream applications.

To date, there are 12 review papers focusing on this field, discussing existing wetland inventory datasets \cite{hu2017global}, supported downstream applications \cite{rebelo2009remote, adam2010multispectral}, classification methods \cite{mirmazloumi2021status}, and remote sensing data for wetland mapping \cite{jafarzadeh2022remote}. We summarize them in Table \ref{tab:SOTAreview}. We can tell that all of them lack of a thorough and in-depth understanding of wetland classification and mapping. 

\begin{table*}[t]
    \centering
    \renewcommand{\arraystretch}{1.1}
    \caption{\label{tab:SOTAreview}Summary of existing remote sensing of wetland classification and mapping-related reviews. TIP denotes Traditional image processing-based methods, ML denotes traditional machine learning-based methods, DL denotes deep learning-based methods. AUX denotes the auxiliary data. Applications denote the downstream applications such as methane emission estimation, flood regulation, etc. } 
    \begin{tabular}{ccccccccccc}
    \hline
       \multirow{2}*{Publicaitons} &  \multicolumn{3}{c}{Reviewed Methods} &  \multicolumn{3}{c}{Reviewed Data} & \multirow{2}*{Applications}& \multirow{2}*{Driving Factors} & \multirow{2}*{Inventories} & \multirow{2}*{Range}  \\  \cline{2-7}
       & TIP & ML & DL & Optical & SAR & AUX \\
       \hline
       \cite{gellman1991forested} & $\checkmark$ & $\times$ & $\times$ & $\times$ & $\times$ & $\times$ & $\checkmark$ & $\times$ & $\checkmark$ & North America \\
       \cite{rebelo2009remote} & $\times$ & $\times$ & $\times$ & $\times$ & $\times$ & $\times$ & $\checkmark$ & $\times$ & $\checkmark$ & Global \\
       \cite{adam2010multispectral} & $\checkmark$ & $\times$ & $\times$ & $\checkmark$ & $\times$ & $\times$ & $\checkmark$ & $\times$ & $\times$ & Global \\
       \cite{amler2015definitions} & $\times$ & $\times$ & $\times$ & $\times$ & $\times$ & $\times$ & $\checkmark$ & $\times$ & $\checkmark$ & East Africa \\
       \cite{hu2017global} & $\checkmark$ & $\times$ & $\times$ & $\checkmark$ & $\times$ & $\times$ & $\times$ & $\times$ & $\checkmark$ & Global \\
       \cite{mahdavi2018remote} & $\checkmark$ & $\checkmark$ & $\times$ & $\checkmark$ & $\checkmark$ & $\times$ & $\times$ & $\times$ & $\checkmark$ & Global \\
       \cite{mahdianpari2020meta} & $\checkmark$ & $\checkmark$ & $\times$ & $\checkmark$ & $\checkmark$ & $\checkmark$ & $\times$ & $\times$ & $\times$ & North America \\
       \cite{adeli2020wetland} & $\times$ & $\times$ & $\times$ & $\times$ & $\times$ & $\checkmark$ & $\times$ & $\times$ & $\times$ & Global \\
       \cite{mirmazloumi2021status} & $\checkmark$ & $\checkmark$ & $\times$ & $\checkmark$ & $\checkmark$ & $\checkmark$ & $\times$ & $\times$ & $\times$ & Canada \\
       \cite{jafarzadeh2022remote} & $\checkmark$ & $\checkmark$ & $\times$ & $\checkmark$ & $\checkmark$ & $\checkmark$ & $\times$ & $\times$ & $\times$ & Global \\
       \cite{demarquet2023long} & $\times$ & $\times$ & $\times$ & $\checkmark$ & $\times$ & $\times$ & $\checkmark$ & $\checkmark$ & $\times$ & Global \\
       \cite{abdelmajeed2023cloud} & $\times$ & $\times$ & $\times$ & $\times$ & $\times$ & $\times$ & $\checkmark$ & $\times$ & $\times$ & Global \\
\hline
\rowcolor[HTML]{FAE3E1} 
        \textbf{Ours} & $\checkmark$ & $\checkmark$ & $\checkmark$ & $\checkmark$ & $\checkmark$ & $\checkmark$ & $\checkmark$ & $\checkmark$ & $\checkmark$ & Global \\ 
\hline
    \end{tabular}
\end{table*}

In the current era, characterized by the overwhelming influence of artificial intelligence and big data following decades of transformative evolutionary paradigms, it is imperative to re-recognize the field of wetland classification and mapping. We must pose and address the following fundamental questions:
\begin{itemize}
    \item What is the scientific importance of wetland classification and mapping regarding the scientific support for ecological protection and assessment?
    \item What kinds of data and methods have been used in wetland classification and mapping, especially based on the intersections of remote sensing big data era and AI-driven era?
    \item What critical insights into wetland changes and driving factors can be derived from wetland classification and mapping research?
    \item What is the current research paradigm and the limitations?
    \item In face of global climate change and technological innovation, what are the challenges and future opportunities for wetland classification and mapping?
\end{itemize}

In consideration of the questions that have been identified, this review aims to offer a comprehensive understanding of wetland classification and mapping from multiple perspectives. The review has three main features: a meta-analysis approach designed to systematically integrate and evaluate prevailing trends and knowledge; a thorough review of scientific importance, key data, methods; and an in-depth discussion regarding current paradigm, limitations and the future of wetland classification and mapping. The central focus encompasses the past, present, and future, further supplemented by various instrumental components such as the driving factors influencing wetland changes and inventory products. Positioned at a pivotal moment in the evolution of the research paradigm, this review aspires to elucidate the development and near future within the wetland classification and mapping field, thereby imparting essential knowledge and concepts for readers to think about and find potential near-and-far future directions.

The rest of this review is organized as follows. We present the meta-analysis of related literature in Sec. \ref{sec:meta}. Following that, we conduct a thorough review in Sec. \ref{sec:review}. After that, we make an in-depth discussion in Sec. \ref{sec:discussion}. We envision our promising prospects on the domain in Sec. \ref{sec:pros}. Finally, we conclude this review in Sec. \ref{sec:conclu}.

\section{Meta-analysis}\label{sec:meta}
For many years, wetlands have drawn the interest of researchers in remote sensing and environmental studies. This section presents a comprehensive meta-analysis of publications related to wetland classification and mapping using remote sensing, covering the period since the 1970s.

\subsection{Overall trend}
The development trends of wetland classification and mapping in remote sensing can be summarized from the following five aspects.

\begin{figure*}[t]
    \centering
    \includegraphics[width=1.0\linewidth]{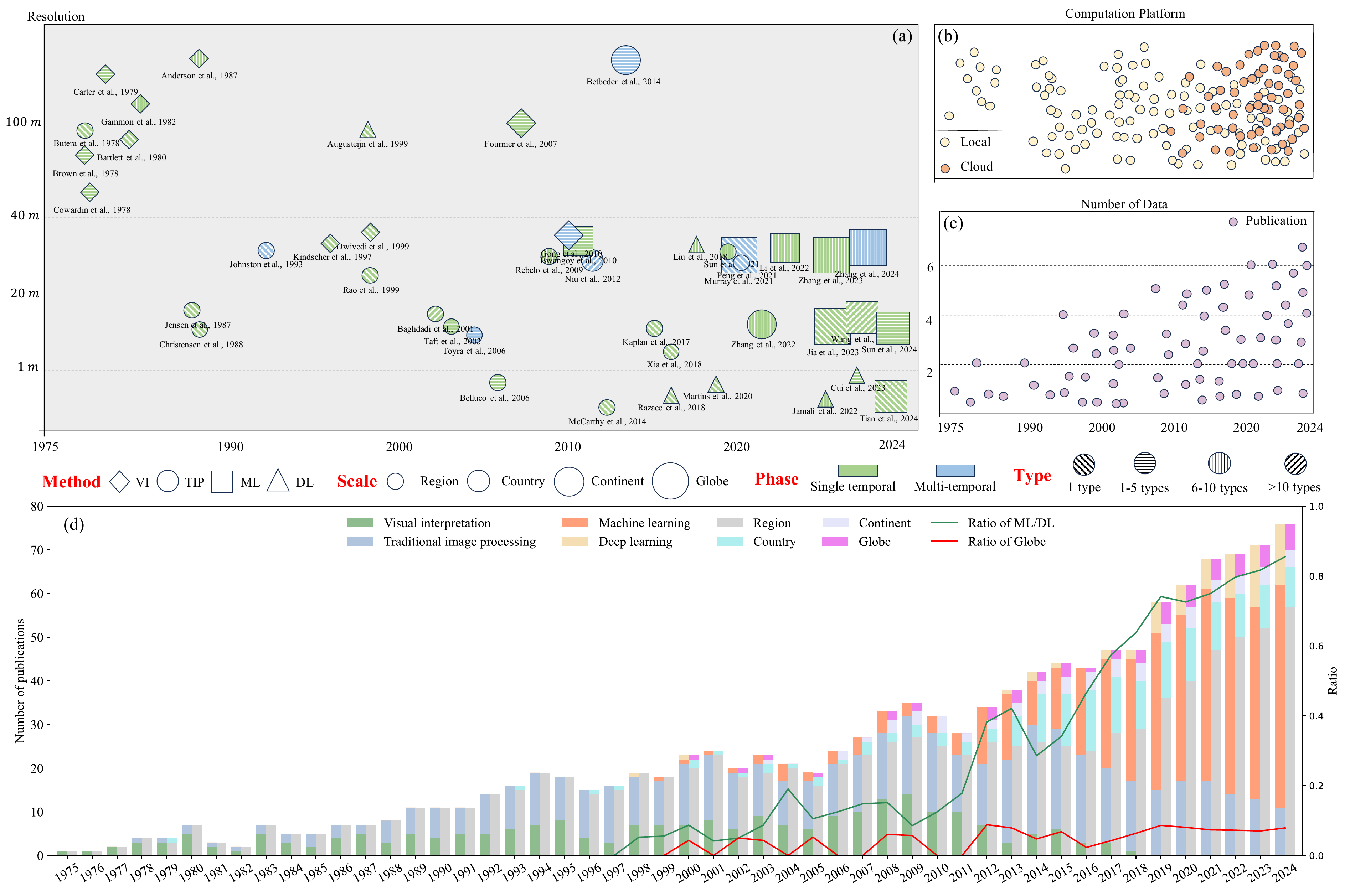}
    \vspace{-1.5em}
    \caption{The overall trend of the wetland classification and mapping from some typical publications since the 1970s. (a) The data resolution, method, scale, phase, and type of typical publications across years. (b) The computation platform choice of typical publications across years. (c) The number of data used for each of the typical publications across years. (d) The statistics of methods and scales in wetland classification and mapping-related publications.}
    \label{fig:trend}
    \vspace{-1.5em}
\end{figure*}

\textbf{Data.} The advancement of remote sensing technology has led to continuous improvements in spatial, temporal, spectral, and radiometric resolutions, resulting in a diverse range of data types and extensive data availability. Research data has evolved from low resolutions to high resolutions (Fig. \ref{fig:trend} (a)) \cite{ tamura1998extraction,di2018land, martins2020deep}, from local processing to cloud-based computation (Fig. \ref{fig:trend} (b)) \cite{kulawardhana2007evaluation, gong2010china, abdelmajeed2023cloud}, from single sources to multiple sources (Fig. \ref{fig:trend} (c)) \cite{niu2012mapping, hosseiny2021wetnet, wang2023wetland}

\textbf{Method.} The growth of diverse remote sensing data has led to intelligent information processing methods, evolving from visual interpretation and manual delineation to semi-automatic and automatic classification technologies. Traditional processing gradually fails to meet accuracy and efficiency needs. Machine learning and deep learning are becoming more and more crucial for wetland classification and mapping. (Fig. \ref{fig:trend} (a)) \cite{delancey2019comparing, hosseiny2021wetnet}.

\textbf{Classification system.} Wetland remote sensing has evolved from simple classification to a more complex, refined system. Initially, it focused on basic wetland types \cite{gao1996ndwi, lowell1998evaluation, gong2013finer}, but ongoing research has expanded the classification system to a maximum of 14 types (Fig. \ref{fig:trend} (a)) \cite{wang2023wetland,mao2024trajectory}.

\textbf{Scale.} Research is shifting to large-scale (national, continental, global) projects \cite{pekel2016high, murray2019global, zhang2024global}. Initially focused on local wetland mapping, the rise in demand for global environmental research and advancements in data processing has led to a trend in large-scale wetland remote sensing. However, local research remains vital. Despite a focus on large-scale studies, in-depth research on ecologically significant or threatened local wetlands continues (Fig. \ref{fig:trend} (a)) \cite{hess2003dual, bwangoy2010wetland, chen2018dynamic, mccarthy2015improved}.

\textbf{Phase.} Research has paid more attention to long time-series wetland mapping to reveal the wetland variations and their correlations with climate changes and human activities \cite{mao2024trajectory, zhang2024global, wang2024interannual}. However, short-term but high-frequency monitoring remains vital for wetlands facing rapid changes or disturbances. During floods or pollution, high-frequency data from Sentinel and UAVs capture wetland dynamics, enhancing the understanding of their characteristics alongside long-term studies (Fig. \ref{fig:trend} (a)) \cite{muro2019mapping}. 

\begin{figure}[h]
    \centering
    \includegraphics[width=1.0\linewidth]{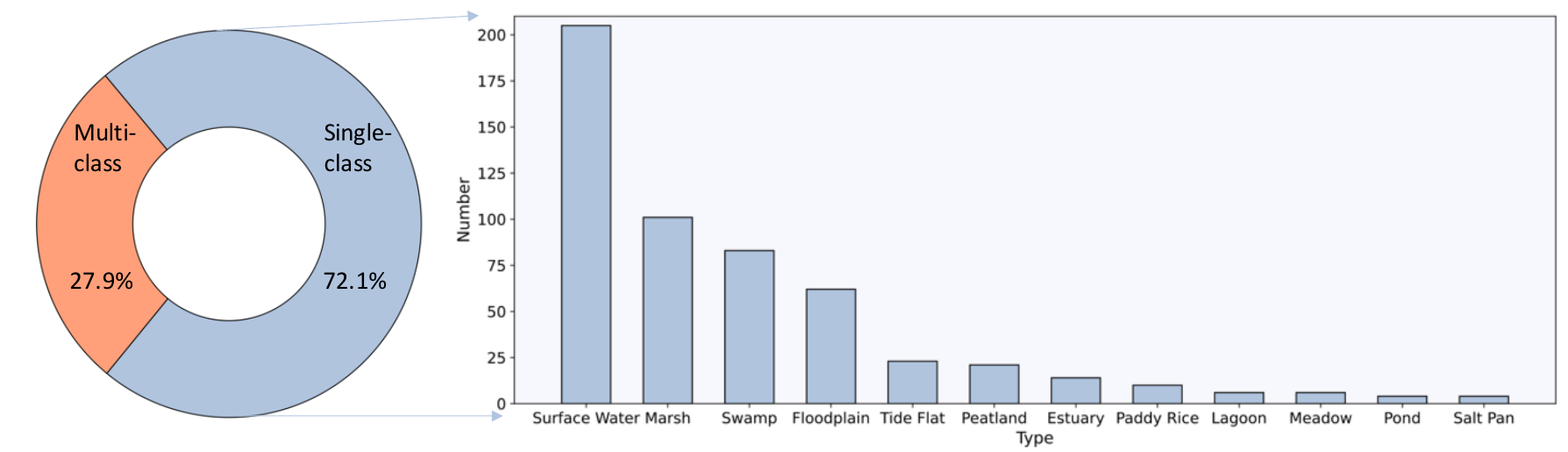}
    \vspace{-1.5em}
    \caption{The statistics of wetland types in wetland classification and mapping publications. Note that papers that treat all wetland types as a single category are excluded.}
    \label{fig:type}
    \vspace{-1.5em}
\end{figure}

\subsection{Quantitative analysis}
\textbf{Wetland types.} Fig. \ref{fig:type} displays the statistics of wetland types in wetland classification and mapping-related publications. It reveals a notable disparity in research categorization, with research on multi-class wetland classification and mapping accounting for a mere 27.9\%, which is approximately one-third of the research on single-class wetland classification and mapping at 72.1\%. This indicates a significant skew towards single-class studies in the wetland research domain. It’s clear that scholars show a strong interest in surface water, with about 210 papers focused on this area. This number significantly surpasses that of other wetland types, like marsh, which has around 100 papers, and swamp, with nearly 80 papers. On the other hand, some other types, such as meadow, pond, and salt pan, receive much less attention, each with fewer than 25 papers recorded. Note that there are over 270 papers that treat all wetland types as a single category, and these papers are excluded from this statistical analysis. 

\textbf{Scale.} Fig. \ref{fig:trend} (d) illustrates the scope of wetland classification and mapping-related publications from the 1970s to the present. This scope is categorized into four distinct levels: region, country, continent, and globe. It is evident that prior to 1993, all publications predominantly focused on typical regions, with the exception of the National Wetland Inventory (NWI) by U.S. Fish \& Wildlife Service \cite{wilen1995us}, which was initiated in 1979 yet completed much later. The first comprehensive global research study was conducted in 2000 \cite{loveland2000development}, which mapped permanent wetlands and water bodies utilizing AVHRR images with a resolution of 1 km. Subsequently, an increase in large-scale research initiatives emerged, attributable to advancements in data processing and observation capabilities. For instance, the number of country-scale studies has experienced a steady increase since 2013. Additionally, it can be observed that since 2009, the proportion of global research has consistently maintained around 10\% of all wetland classification and mapping-related publications. Despite this ratio remaining stable, the absolute number of global publications continues to rise in conjunction with the overall increase in publication volume. 

\begin{figure}[h]
    \centering
    \includegraphics[width=1.0\linewidth]{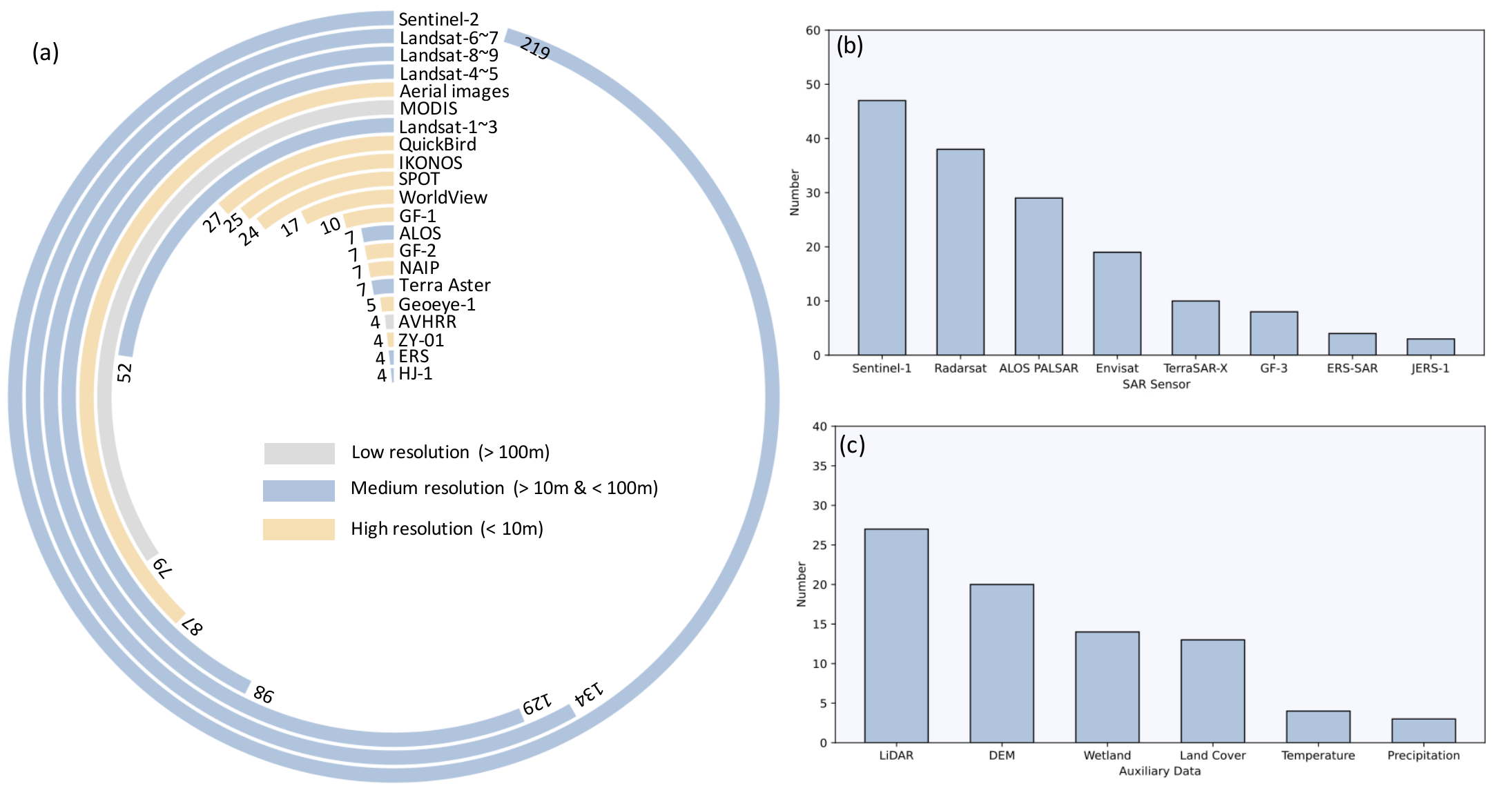}
    \vspace{-1.5em}
    \caption{The number of data types used in wetland classification and mapping-related publications. (a) Optical sensors. (b) SAR sensors. (c) Auxiliary data.}
    \label{fig:data}
    \vspace{0em}
\end{figure}

\textbf{Data.} As shown in Fig. \ref{fig:data} (a), the data utilized in wetland classification and mapping-related research can be categorized into three main groups: optical sensors, Synthetic Aperture Radar (SAR) sensors, and auxiliary data. For optical sensors, the data is further classified by resolution into low-resolution (>100m), medium-resolution (>10m and <100m), and high-resolution (<10m). Low-resolution data, while used less frequently and represented by fewer sources, remains a valuable tool for broad-scale wetland classification and mapping applications. Among them, MODIS stands out as the most frequently used low-resolution dataset, employed 79 times. Despite their efficiency and convenience in large-area monitoring, low-resolution datasets often lack the spatial detail needed to accurately capture wetland features. This results in lower accuracy and a more generalized understanding of wetland conditions, particularly in regions where fine-scale distinctions are necessary. The medium-resolution category, which includes the Landsat series, Sentinel-2, and ALOS, represents a substantial portion of wetland classification and mapping-related research. Landsat has been utilized the most, with 358 instances of use, followed by Sentinel-2 with 219 instances. The extensive accessibility, long-term coverage, and global availability of these datasets make them highly popular for monitoring both long-term wetland changes and spatially extensive wetland areas. Medium-resolution data is sufficient for the identification of various wetland types, including marshes, swamps, and tidal flats, and provides a more refined and explicit understanding of wetland dynamics. High-resolution data, predominantly sourced from commercial satellites or aerial imagery, is typically used in small-scale studies where the need for precise, fine-grained classifications is paramount. Given the large volume of high-resolution data, these datasets are employed primarily for localized wetland mapping, where detailed ecological features and small-scale variations are crucial for accurate classification.

Fig. \ref{fig:data} (b) shows Synthetic Aperture Radar (SAR) data, which plays a crucial role in monitoring wetland dynamics, particularly in areas where cloud cover hinders optical observations. Compared to optical data, SAR data sources are fewer and less frequently used. Among them, Sentinel-1 is the most commonly utilized, with 47 instances of use. Fig. \ref{fig:data} illustrates the auxiliary data. This category encompasses a variety of supplementary datasets that enhance the accuracy and detail of wetland mapping by providing critical contextual information. The most used is LiDAR (27 times) and DEM (20 times), both providing the precise topography and surface elevation of wetlands. Existing wetland inventories and land cover data are also used in some research for offering historical data on wetland extent, which can be used for comparison, training, validation in current mapping efforts. 


\textbf{Method.} Fig. \ref{fig:trend} (d) further illustrates the trends and statistics of the development of methodologies in wetland classification and mapping-related publications. Before the year 2000, the majority of research employed visual interpretation and conventional image processing techniques for wetland classification and mapping. However, since the work of \cite{augusteijn1998wetland}, which first utilized neural networks for wetland classification, there has been a marked increase in the incorporation of machine learning and deep learning approaches within wetland classification and mapping-related publications, which now constitute over 80\% of this field. This shift reflects the broader transition in remote sensing from rule-based and expert-driven approaches to data-driven and automated classification frameworks. Traditional methods, such as index-based classification and object-based image analysis, were widely employed in early wetland classification studies due to their interpretability and relatively simple computational requirements. However, their reliance on predefined statistical assumptions and handcrafted feature selection often limited their ability to generalize across diverse wetland environments, particularly in heterogeneous landscapes where spectral confusion is prevalent. The increasing availability of multi-source, multi-temporal, and high-resolution remote sensing data has accelerated the adoption of machine learning and deep learning techniques due to their robustness in handling high-dimensional datasets and their ability to model nonlinear relationships. More recently, deep learning architectures—such as convolutional neural networks (CNNs), generative adversarial networks (GANs), and transformer-based models—have further revolutionized wetland mapping by automatically learning hierarchical spatial-spectral features, thereby reducing the need for handcrafted feature engineering. 
\begin{figure*}[h]
    \centering
    \includegraphics[width=1.0\linewidth]{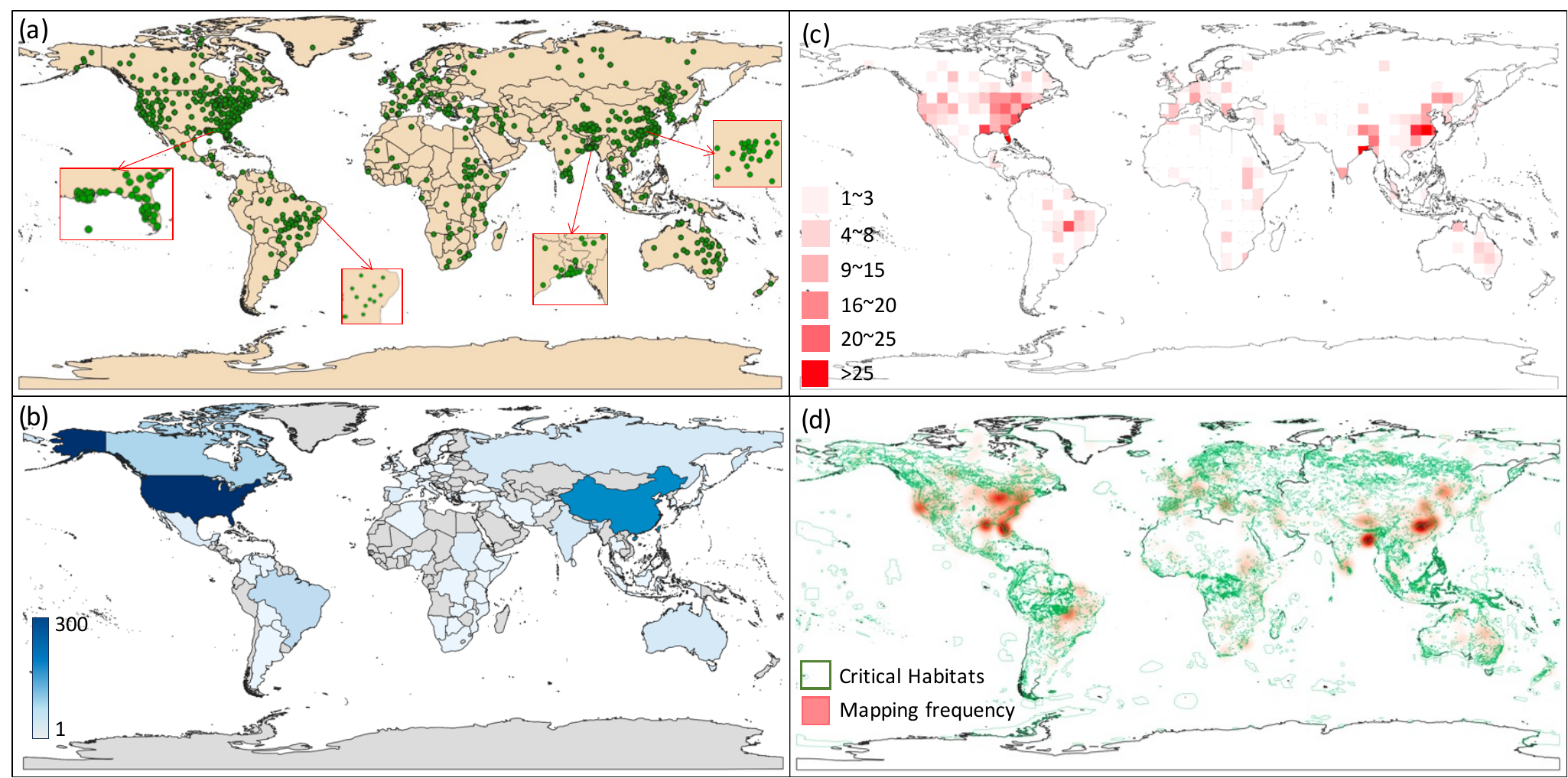}
    \caption{The number and distribution of study sites around the world according to our database. (a) The study site distributions. (b) The number of publications at country scale. (c) The density map of study sites. (d) The density map of study sites overlaid by the critical habitat map.}
    \label{fig:sites}
    \vspace{0em}
\end{figure*}

\textbf{Study sites.} Fig. \ref{fig:sites} presents the quantitative data of research sites involved in wetland classification and mapping-related publications. Fig. \ref{fig:sites} (a) shows the detailed research sites since 1975. Fig. \ref{fig:sites} (b) presents the number of publications at a country scale. The United States stands out with more than 280 recorded instances, followed by China with over 150 instances, and Canada with 89. Regarding the spatial distribution of wetlands, most study areas are located in North America, East Asia, and Europe. Additionally, countries in the Southern Hemisphere, like Brazil and Australia, gain attention for their unique wetland types, which play a significant role in global climate change. Fig. \ref{fig:sites} (c) shows the density map of the research sites. The east of the United States (e.g., Great Lakes and Florida), the east of China (e.g., the Middle-lower Yangtze Plain), the Amazon Plain and Pantanal Wetlands, and the Bay of Bengal are the research hotspots. Fig. \ref{fig:sites} (d) shows the overlay of the density map and the critical habitat map \cite{martin2015global,brauneder2018global,UNEP-WCMC2017}. We can see there is a critical mismatch between the biodiversity protection regions and the high-mapping-frequency regions. For example, Africa, despite its rich research potential and extensive wetlands and habitats, has a notably lower number of research instances due to its complex topography, frequent cloud cover, limited field study opportunities, poor photographic conditions and few local research teams. This highlights the emerging need for global-scale wetland mapping efforts to pay attention to all wetland ecosystems equally and comprehensively, ensuring representation across diverse wetland types, geographic regions, and ecological conditions.

\section{Thorough Review}\label{sec:review}

\begin{table*}[t]
\caption{Summary of typical data sources in wetland classification and mapping.}
\label{tab:method}
\renewcommand{\arraystretch}{1.5}
\resizebox{\linewidth}{!}{
\begin{tabular}{ccccccc}
\hline
Data Type                       & Sub-Category                       & Name         & Spatial Resolution & Number of Band & Sapnning Time & Revisit Time     \\ \hline
\multirow{10}{*}{Optical Data}  & \multirow{4}{*}{High-resolution}   & CASI-1500    & \textless \ 0.1m     & 144            & -             & -                \\
                                &                                    & IKONOS       & 1 m/4 m            & 4              & 1999-2015     & 3 days           \\
                                &                                    & WorldView    & 0.3 m-1.8 m        & 4-8            & 2007-Now      & \textless \ 2 days \\
                                &                                    & Gaofen-2     & 0.8 m/3.2 m        & 4              & 2014-Now      & 5 days         \\ \cline{2-7}
                                & \multirow{4}{*}{Medium-resolution} & Sentinel-2   & 10 m               & 13             & 2015-Now      & 5-10 days        \\
                                &                                    & SPOT         & 10 m/ 20 m         & 1-4            & 1986-2013     & 4-5 days         \\
                                &                                    & Landsat 7-9  & 30 m               & 8-11           & 1999-Now      & 16 days          \\
                                &                                    & ALOS         & 2.5 m/10 m         & 4              & 2006-2011     & 2 days           \\ \cline{2-7}
                                & \multirow{2}{*}{Low-resolution}    & MODIS        & 250 m-1000 m       & 36             & 1999-Now      & 1 day            \\
                                &                                    & AVHRR        & 1000 m             & 6              & 1988-Now      & 1 day            \\ \hline
\multirow{5}{*}{SAR Data}       & \multirow{5}{*}{-}                 & RADARSAT-1   & 8-100 m            & C              & 1995-2013     & 24 days          \\
                                &                                    & TerraSAR-X & 0.25-40 m          & X              & 2007-Now     & 11 days          \\
                                &                                    & Sentinel-1   & 5-40 m             & C              & 2014-Now      & 12 days          \\
                                &                                    & ALOS-PALSAR  & 1-100 m            & L              & 2014-Now      & 14 days          \\
                                &                                    & Gaofen-3     & 1-100 m            & C              & 2016-Now      & \textless \ 3 days \\ \hline
\multirow{5}{*}{Auxiliary Data} & LiDAR                              & ICESat       & 70 m               & 1              & 2003-Now      & -                \\ \cline{2-7}
                                & DEM                                & SRTM         & 30 m/90 m          & 1              & 2000          & -                \\ \cline{2-7}
                                & Land Cover                         & FROM-GLC 10  & 10 m               & 1              & 2017          & -                \\ \cline{2-7}
                                & Wetland                            & GLWD-v2      & 500 m              & 1              & -             & -                \\ \cline{2-7}
                                & Climate                            & ERA5-Daily   & 30 km              & 9              & 1979-Now      & 1 day            \\ \hline
\end{tabular}}
\end{table*}

\subsection{Wetland features}
Wetlands present unique challenges and opportunities for remote sensing due to their complex spectral, spatial, and scattering characteristics.

Spectrally, wetland surfaces exhibit substantial complexity driven primarily by high spatial and spectral variability, steep environmental gradients, and closely overlapping spectral signatures among different vegetation types. Solar radiation interacting with wetland environments undergoes reflection, absorption, and transmission processes, resulting in spectral signatures that are significantly more intricate compared to terrestrial land covers. Vegetation canopies in wetlands share similar biochemical compositions, leading to overlapping absorption features. The presence of underlying substrates such as soils, standing water, and high atmospheric moisture further complicates spectral reflectance, especially in near-infrared (NIR) and mid-infrared (MIR) wavelengths, diminishing the effectiveness of vegetation discrimination. Although visible wavelengths primarily delineate open water areas, NIR wavelengths are particularly valuable in identifying water-land transition zones. Additionally, MIR wavelengths, sensitive to moisture variations in both vegetation and soils, provide critical insights into wetland hydrological conditions.

Spatially, relying solely on spectral data often results in misclassification due to phenomena like spectral confusion, where different objects exhibit identical spectral signatures or the same objects display diverse spectral characteristics. The spatial arrangement and texture patterns of pixels thus become essential distinguishing factors. Wetlands frequently demonstrate distinct spatial configurations, such as coastal saline lakes manifesting as polygonal shallow water bodies isolated by islands or reefs, interconnected to oceans through narrow channels. Beyond these straightforward water features, wetlands are characterized by complex mosaics involving water bodies, shorelines, and diverse aquatic vegetation communities, complicating spatial analyses.

The scattering properties of wetlands offer complementary insights crucial for accurate characterization, especially through synthetic aperture radar (SAR) data. SAR effectively captures the three-dimensional structure of wetlands by utilizing the interaction between microwave signals and the wetland surface. Static water surfaces typically generate weak backscatter signals, appearing dark in SAR imagery, while flooded vegetation, exhibiting double-bounce scattering, produces bright signals. Non-flooded vegetation tends to display intermediate brightness through volume scattering. 

\subsection{Data}\label{sec:data}
The exponential growth of Earth observation platforms has revolutionized data availability for wetland classification and mapping , yet simultaneously introduced critical questions about how to effectively harness multi-source geospatial data. While early wetland classification and mapping predominantly relied on single-source remote sensing data, now more and more approaches reconcile multi-source distinct yet complementary data categories: remote sensing optical data, remote sensing SAR data, and other auxiliary data (e.g., LiDAR, DEM, land cover, and wetland inventories). Each data type captures unique facets of wetland ecosystems but presents specific technical challenges for systematic integration.

\subsubsection{Optical data}
Remote sensing optical data capture visible and infrared light, offering spectral information crucial for wetland studies. The category includes various satellites, and we divide them based on the spatial resolution.

\textbf{High-resolution.} High-resolution optical data play a crucial role in extracting the detailed spatial characteristics of wetlands. These data sources, with their enhanced geometric and textural details, enable precise delineation of wetland boundaries, identification of micro-topographic features, and characterization of vegetation patterns at fine scales. Among the most prominent high-resolution systems are the American IKONOS and WorldView series satellites, which provide global coverage of panchromatic and multispectral imagery with resolutions over 1 meter. The subsequent WorldView series, Gaofen series, further enhanced these capabilities with improved spectral resolution (8-band and 16-band multispectral sensors, respectively) and advanced radiometric performance. These high-resolution datasets have found extensive applications in wetland research, particularly in: Detailed habitat mapping and classification of wetland vegetation communities \cite{hladik2013salt}; Monitoring of fine-scale changes in wetland morphology and hydrology \cite{zhou2010monitoring}; Identification and mapping of invasive species distribution patterns \cite{laba2008mapping}; Validation and refinement of wetland maps derived from medium-resolution sensors \cite{han2015four}.


\textbf{Medium-resolution.} Medium-resolution optical data have emerged as a powerful tool for large-scale wetland monitoring, offering an optimal balance between spatial detail and temporal frequency. These datasets, typically with resolutions ranging from 10 to 100 meters, are particularly well-suited for regional and global wetland assessments due to their extensive spatial coverage and regular revisit cycles. Among the most prominent sources is the Landsat series, which has provided nearly five decades of continuous multispectral observations with 30-meter spatial resolution and 16-day revisit cycle, making it the cornerstone of long-term wetland monitoring and change detection studies \cite{pekel2016high, wang2024interannual}. A significant enhancement to medium-resolution monitoring capabilities came with the European Space Agency's Sentinel-2 mission, which offers a 5-day global revisit frequency when combined with its twin satellites, Sentinel-2A and Sentinel-2B. With spatial resolutions of 10 meters (visible and near-infrared bands) and 20 meters (shortwave infrared bands), Sentinel-2 data provide improved spectral resolution and temporal frequency compared to Landsat, particularly valuable for capturing rapid wetland changes such as seasonal inundation patterns and vegetation phenology \cite{liu2022precise}. The synergy between Landsat and Sentinel-2 datasets, facilitated by harmonization efforts like the Harmonized Landsat and Sentinel-2 (HLS) project, has significantly enhanced the capacity for high-frequency wetland monitoring.

The unique advantages of medium-resolution data make them indispensable for various wetland applications. Their high temporal frequency enables the detection of seasonal and interannual variations in wetland extent and condition, while their broad spatial coverage supports continental-scale assessments and cross-border ecosystem studies. These datasets have been instrumental in global wetland mapping initiatives, such as the Ramsar Convention's wetland inventory updates, and have played a crucial role in monitoring wetland responses to climate change and human activities \cite{murray2019global}. 

\textbf{Low-resolution.} Low-resolution optical data, characterized by spatial resolutions ranging from 100 meters to several kilometers, serve as important tools for global-scale wetland classification and mapping studies. These datasets are particularly valuable for capturing broad-scale wetland dynamics, assessing ecosystem functions, and analyzing long-term environmental changes. Among the most widely used low-resolution sensors are  MODIS and AVHRR, which have revolutionized our ability to observe Earth's surface at continental and global scales.

The unique advantages of low-resolution data lie in their ability to provide consistent, long-term observations at global scales, enabling researchers to address fundamental questions about wetland distribution, ecological functions, and responses to global environmental change. These datasets have been instrumental in large-scale or high-frequency wetland studies, such as mapping the extent of global large lakes \cite{lehner2004development}, dynamic wetland inundation mapping \cite{chen2013evaluation}. However, the coarse spatial resolution of these sensors limits their utility for studying small or fragmented wetlands, necessitating integration with higher-resolution datasets for comprehensive wetland assessments. Recent advances in data fusion techniques and machine learning algorithms have significantly enhanced the value of low-resolution data by enabling their combination with medium- and high-resolution observations \cite{chen2024global,chen2023robot}, opening new possibilities for multi-scale wetland monitoring and analysis.

\subsubsection{SAR Data}
Wetland ecosystems have a three-dimensional structure with perennial surface water at varying depths, supporting diverse underwater, floating, and emergent vegetation. This structure can create spectral and spatial ambiguities compared to similar land surfaces. Thus, the scattering properties of wetlands are vital for remote sensing observations. Synthetic Aperture Radar (SAR) is an active microwave remote sensing system that captures backscattering signals essential for wetland studies. These signals respond to surface roughness, vegetation structure, and dielectric properties, making SAR crucial for wetland characterization. Moreover, SAR's ability to function in all-weather and day-night conditions addresses the challenges of cloud cover and solar illumination that affect optical remote sensing. This capability allows it to monitor the phenological cycles of wetland vegetation and dynamic water levels over time. SAR's unique imaging mechanism, along with its sensitivity to structural and moisture properties, complements the information gained from optical sensors. 
Among all SAR data, Sentinel-1 with C-band SAR, offers high temporal frequency and is widely used for monitoring complex wetland extent and distributions. A synergic use of Sentinel-1 and Sentinel-2 imagery for complex wetland classification using machine learning techniques \cite{wang2023wetland, liu2022precise} demonstrated its effectiveness in dynamic environments. TerraSAR-X provides high-resolution X-band with a longer wavelength and is used for mangrove monitoring. Gaofen-3, China's first civilian C-band SAR satellite, offers high-resolution imaging capabilities (up to 1 meter in spotlight mode) and multiple polarization modes (quad-pol, dual-pol, and compact-pol). This versatility enables detailed mapping of wetland microtopography and fine-scale vegetation patterns, particularly in complex wetland landscapes such as coastal marshes and riverine floodplains \cite{zhang2021coastal}.

\subsubsection{Auxiliary data}
Auxiliary data complement remote sensing by providing additional context and improving classification accuracy, including LiDAR, DEM, land cover, wetland inventories and climate data.

\textbf{LiDAR}
Light Detection and Ranging (LiDAR) data have emerged as a transformative tool in wetland mapping, offering unparalleled capabilities for capturing the three-dimensional structure of wetland ecosystems. Unlike traditional optical or SAR remote sensing, LiDAR, such as ICESat, provides direct measurements of surface elevation and vegetation structure through laser pulse returns, enabling precise characterization of wetland microtopography and canopy architecture. This high-resolution vertical information is particularly valuable for delineating subtle hydrological gradients, identifying small-scale drainage features, and mapping vegetation strata in complex wetland environments \cite{huang2014wetland}.

\textbf{DEM.} Digital Elevation Models (DEMs) serve as foundational datasets in wetland mapping, providing critical topographic information that underpins hydrological modeling and ecosystem characterization. As a representation of the Earth's surface elevation, DEMs enable the identification of subtle terrain features that control water movement, accumulation, and drainage—key processes governing wetland formation and distribution. The integration of DEMs with other geospatial data has revolutionized our ability to predict wetland locations, delineate their boundaries, and understand their hydrological connectivity at landscape scales \cite{bwangoy2010wetland}. 

\textbf{Land cover.} Land cover data, derived from remote sensing, classify surface types including wetlands, providing context for classification. Provided with prior wetland distributions, land cover data can enhance understanding of the broader landscape, aiding in distinguishing wetland from other land cover types, and are crucial for comprehensive mapping \cite{gong2013finer}. 

\textbf{Wetland inventories.} 
Wetland inventories are databases or maps providing reference information on wetland locations and types. Also as a prior knowledge of the existing wetland distribution, types and extents, they are used for complementary training sample generation, validation and as a source of ground truth data, providing historical context for mapping efforts. These inventories are vital for assessing new mapping efforts' accuracy and providing historical context, supporting conservation and management decisions \cite{zhang2024global}.

\textbf{Climate data.} Climate data, especially temperature and precipitation data, are complementary for precise wetland mapping, governing hydrological and ecological processes like evapotranspiration, water balance, and flood frequency. Temperature influences wetland persistence and boundary shifts, particularly in climate-sensitive regions, while precipitation patterns shape wetland characteristics through surface water levels and seasonal inundation. Integrated with remote sensing data, climate data enhances hydrological modeling and climate change assessments \cite{jafarzadeh2024enhancing}.

\subsection{Method}\label{sec:method}
Traditional methods for wetland classification and mapping largely rely on domain experts manually interpreting aerial photographs or satellite images to produce thematic wetland maps. Although geospatial techniques—such as feature vectorization, symbolization, and cartographic composition—have become increasingly mature \cite{yao2024vectorization, kraak2020cartography}, the diverse and ever-expanding sources of Earth observation data (e.g., spaceborne, airborne, and ground-based sensors) pose unique challenges for wetland mapping: how can we develop advanced methods to fully utilize the diverse data to map wetlands accurately and profoundly? In particular, wetland ecosystems often exhibit high spatial heterogeneity and temporal variability, making it difficult to quickly and efficiently interpret critical wetland features within these large geospatial datasets. Owing to the complex spatial distribution and fuzzy boundaries of wetlands, there is low discriminability between classes yet high variability within each class. This makes wetland classification—especially the recognition of subtypes—particularly challenging. As summarized in Table \ref{tab:method}, contemporary wetland remote sensing interpretation approaches can be systematically grouped into three main paradigms: traditional image processing methods, machine learning methods, and deep learning methods.

\begin{table*}[h]
\caption{Summary of contemporary wetland classification and mapping approaches.}
\label{tab:method}
\renewcommand{\arraystretch}{1.5}
\begin{tabular}{cclll}
\hline
Method                                        & Sub-category      & \multicolumn{3}{c}{Typical algorithms}                  \\ \hline
\multirow{6}{*}{Traditional image processing} & Pixel feature     & \multicolumn{3}{l}{
\begin{tabular}[c]{@{}l@{}}
NDVI \cite{xu2018comparison, ashok2021monitoring, dong2014mapping, chen2014dynamic}, NDWI \cite{yang2017mapping, gao1996ndwi, ahmed2017analysis}, EVI \cite{dai2020vegetation, zhao2009monitoring, yan2010detecting}, LSWI \cite{zhao2009monitoring, li2015evaluation}
\end{tabular}} \\ \cline{2-5}

& Object analysis   & \multicolumn{3}{l}{
\begin{tabular}[c]{@{}l@{}}
Feature: Perimeter \cite{dronova2015object, ruan2012mapping}, Area \cite{berhane2017comparing}, Compactness \cite{zhou2021object}, Roundness \cite{mao2020national}, SIFT \cite{liu2019evaluating}, Harris\\
Segmentation: Mean-Shift Method \cite{dronova2011object}, Fractal Network Evolution \cite{fu2017comparison, bertassello2018wetlandscape}
\end{tabular}} \\ \cline{2-5}

& Transform         & \multicolumn{3}{l}{
\begin{tabular}[c]{@{}l@{}}
Gabor Wavelet \cite{hatvani2017periodic,li2021urban}, Tasseled Cap Transformation (TCT) \cite{wang2023wetland}, Curvelet Transform \cite{schmitt2013wetland}
\end{tabular}} \\ \cline{2-5}

& Spectral unmixing & \multicolumn{3}{l}{Internal Maximum Volume \cite{hong2021interpretable}, Dependent Component Analysis \cite{na2021wetland}, Least Squares \cite{na2021wetland}} \\ \cline{2-5}

& Sub-pixel & \multicolumn{3}{l}{Spatial attraction models \cite{sidike2019dpen}, genetic algorithms \cite{li2015super}, BP neural networks \cite{li2015super}} \\ \cline{2-5}
& Super-pixel          & \multicolumn{3}{l}{Simple Linear Iterative Clustering \cite{mohseni2023wetland, zeng2024machine}, Simple Non-Iterative Clustering \cite{guo2024mapping, yang2021high}} \\ \hline

Machine learning    & Classifier        & \multicolumn{3}{l}{
\begin{tabular}[c]{@{}l@{}}
Nearest Neighbor Method \cite{na2015mapping, judah2019integration}, Maximum Likelihood Classification \cite{jollineau2008mapping, lane2014improved}, Decision Tree \cite{berhane2018decision}, \\
Support Vector Machine \cite{lin2013remote}, Random Forest \cite{tian2016random, mahdianpari2017random}, K-means \cite{jing2020exploring}
\end{tabular}} \\ \hline

\multirow{5}{*}{Deep learning} 
& CNN               & \multicolumn{3}{l}{CNN family \cite{pouliot2019assessment, rezaee2018deep, li2022evaluation}}   \\ \cline{2-5}
& Region proposal   & \multicolumn{3}{l}{R-CNN family\cite{ma2022identifying, guirado2021mask}, YOLO family} \\ \cline{2-5}
& FCN               & \multicolumn{3}{l}{
\begin{tabular}[c]{@{}l@{}}
FCN family \cite{li2022spectral, pashaei2019fully}, U-Net family \cite{dang2020coastal, li2021mapping}, PSPNet \cite{chouhan2023superpixel}
\end{tabular}} \\ \cline{2-5}

& Transformer       & \multicolumn{3}{l}{ViT family \cite{jamali2022swin, marjani2024cvtnet}} \\ \cline{2-5}
& GAN & \multicolumn{3}{l}{GAN family \cite{he2019wetland, jamali20223dunetgsformer}}
\\ \hline

\end{tabular}
\end{table*}

\subsubsection{Traditional image processing}

Traditional image processing techniques have long served as the backbone for wetland classification and mapping due to their simplicity, interpretability, and relatively low computational cost. 

\textbf{Pixel feature.} Methods such as calculating spectral indices allow for quick discrimination between water, vegetation, and soil \cite{gao1996ndwi, teng2021assessing, hu2018quantitative, acharya2018evaluation}. For example, \cite{teng2021assessing} uses the Normalized Difference Water Index (NDWI) from Sentinel-2 to indicate hydrological conditions, explaining water level and geese distribution. \cite{ashok2021monitoring} uses the Normalized Difference Vegetation Index (NDVI) and NDWI to monitor the dynamic changes of a specific wetland. However, these methods simply rely on the reflectance of remote sensing images, which may be adversely affected by sensor noise and atmospheric disturbances. Besides, They struggle with the high spatial heterogeneity and blurred boundaries typical of wetland ecosystems because significant manual parameter tuning (e.g., static thresholding) is often necessary. 

\textbf{Object analysis.} Object-based classification methods group homogeneous pixels into "objects," combining spectral features \cite{dronova2015object} with spatial attributes such as shape, texture, and size \cite{berhane2017comparing, dingle2015object}. For example, \cite{mui2015object} selected shape, scale and compactness as the object feature for object segmentation. These methods often achieve higher accuracy and more regularized results by leveraging spatial context, making them particularly suitable for wetland remote sensing. However, their reliance on image segmentation and feature extraction adds complexity and potential challenges to the classification process. 

\textbf{Transformation.} Transformation-based methods employ spectral or texture transformations to efficiently capture multi-scale features. Notably, Gabor Wavelet transforms \cite{lee1996image, hatvani2017periodic} have been utilized in wetland mapping for their proficiency in identifying multi-scale and multi-directional textures. These wavelets decompose remote sensing images into frequency and orientation components, effectively characterizing wetlands' spatial patterns, including vegetation, water, and soil. Furthermore, integrating Gabor Wavelet features with spectral indices enhances river-eutrophic pond-wetland classification, as shown in \cite{hatvani2017periodic}. Tasseled Cap Transformation (TCT) converts multispectral data into interpretable components—brightness, greenness, and wetness—to enhance feature separation for wetland mapping analysis. Using TCT, \cite{wang2024interannual} integrated multi-temporal Sentinel-2 data, effectively monitoring seasonal changes in urban wetland and capturing dynamic processes. Curvelet Transform, tailored for multi-scale geometric analysis, is adept at capturing curved edges and directional features, crucial for delineating complex boundaries and textures in wetland systems. As demonstrated in \cite{schmitt2013wetland}, Curvelet Transform significantly enhances edge detection of wetlands in RADARSAT SAR imagery. 

\textbf{Spectral unmixing.} Wetlands act as transitional zones between land and water, resulting in mixed pixels due to complex spatial and spectral traits. Ambiguous boundaries at the water-land interface and the mixing of wetland vegetation and soil spectra exacerbate this issue. These mixed pixels hinder classification accuracy, requiring decomposition for better analysis of wetland distributions at a sub-pixel level. Scholars use spectral unmixing to address spectral mixing, offering insights at the sub-pixel level. It consists of two models: linear and nonlinear mixing models. Nonlinear spectral models more closely align with actual hyperspectral remote sensing conditions but consider radiation quantities among ground objects, which complicates model structure and practical application challenges. Linear models assume that
photons interact with a single material without inter-material
interaction, simplifying the structure and clarity of physical
meaning. For example, Least Squares is one of linear models to decompose mixed pixels into fractional abundances of known spectral endmembers. Given an observed pixel spectrum \(\mathbf{x} \in \mathbb{R}^{d}\), it is modeled as a linear combination of \(M\) endmembers:

\begin{equation}
\mathbf{x} = \sum_{j=1}^{M} a_j \mathbf{e}_j + \mathbf{\epsilon},
\label{eq:lssu_model}
\end{equation}

where \(\mathbf{e}_j \in \mathbb{R}^{d}\) are the spectral endmembers, and \(a_j\) represents the abundance fraction of endmember \(j\), and \(\mathbf{\epsilon}\) is the residual error.
\cite{na2021wetland} applied least squares model on HJ-1A/B imagery to map wetlands, effectively distinguishing between marsh, meadow and open water. Their study demonstrated that unmixing approaches can improve wetland mapping in heterogeneous environments where mixed water and vegetation pixels dominate. However, the dependency on accurate endmember extraction hinders the effectiveness in more complex scenarios.

\textbf{Sub-pixel.} Spectral unmixing models are inadequate in effectively addressing the spatial uncertainties inherent in the mixed pixel problem. Sub-pixel mapping not only quantifies the proportions of endmembers within mixed pixels but also elucidates their spatial distribution. Traditional sub-pixel mapping methodologies can be categorized into two distinct types based on the prior knowledge they incorporate: those founded on spatial correlation assumptions \cite{sidike2019dpen}, those that utilize statistical regression frameworks \cite{li2015super}, and those dependent on hand-crafted regularization filters \cite{li2015super}. \cite{li2015super} has shown the success of subpixel-based methods in accurately mapping wetland inundation in complex spectral-mixing scenarios.

\textbf{Super-pixel.} While sub-pixel-based methods solve the decomposition of spectral mixing, the boundaries cannot be precisely extracted. Super-pixel methods segment an image into perceptually meaningful regions, reducing within-class spectral variability while preserving object boundaries. Simple linear iterative clustering (SLIC) employs k-means clustering in a five-dimensional space (spatial coordinates and spectral values) to iteratively refine super-pixel regions, ensuring compact and homogeneous clusters. \cite{mohseni2023wetland} applied SLIC-based segmentation on Sentinel-1/2 imagery to delineate wetland patches before classification, effectively reducing noise and improving the identification of wetland subclasses such as emergent and submerged vegetation. Simple non-iterative clustering (SNIC) offers a computationally efficient alternative by constructing super-pixels through a region-growing approach, eliminating the need for iterative optimization. \cite{wang2023wetland} utilized SNIC to preprocess Sentinel-2 for large-scale wetland classification, demonstrating that SNIC-super-pixel segmentation enhanced object-based classification accuracy, particularly for fragmented wetland types.


\subsubsection{Traditional Machine learning} Traditional machine learning (ML) techniques have established themselves as indispensable tools for wetland classification and mapping, owing to their robust capacity to model intricate, non-linear relationships and process high-dimensional data. These methodologies have propelled significant advancements in the field by enhancing classification precision, facilitating automated feature extraction, and empowering the scaling ability.

\textbf{Supervised Learning.} Supervised learning methods, such as support vector machines (SVM), random forests (RF), and decision trees (DT). SVMs are effective in handling high-dimensional data and non-linear relationships, making them suitable for wetland classification. However, SVMs require careful parameter tuning (e.g., kernel selection) and may struggle with large datasets due to their computational complexity. RFs are ensemble methods that combine multiple decision trees to improve classification accuracy and reduce overfitting. Given a dataset \(D = \{(\mathbf{x}_i, y_i)\}_{i=1}^{N}\), where \(\mathbf{x}_i\) represents feature vectors and \(y_i\) is the corresponding class label, the prediction of a single decision tree is given by:

\begin{equation}
h_t(\mathbf{x}) = \arg\max_{c} \sum_{i=1}^{N} w_i^{(t)} \mathbb{I}(y_i = c),
\label{eq:decision_tree}
\end{equation}

where \(h_t(\mathbf{x})\) is the predicted class by the \(t\)-th tree, and \(w_i^{(t)}\) is the weight of sample \(i\) in the training process, and \(\mathbb{I}(\cdot)\) is an indicator function that returns 1 if the condition holds, otherwise 0. The final Random Forest prediction is obtained by aggregating the outputs of \(T\) decision trees using majority voting. They are robust to noise and can handle both spectral and spatial features, making them ideal for wetland mapping. \cite{zhang2024global} demonstrated the effectiveness of RF in mapping large-scale wetlands by integrating spectral indices and auxiliary data. Despite their advantages, RFs may produce less interpretable results compared to simpler models like Decision Trees. DTs are intuitive and easy to interpret, making them suitable for small-scale wetland studies. However, they are prone to overfitting and may not perform well on highly complex datasets.  

\textbf{Unsupervised Learning.} Unsupervised learning methods, such as clustering algorithms, do not require labeled data and are useful for exploratory analysis and identifying patterns in wetland ecosystems.  K-Means is a simple and efficient algorithm for grouping pixels based on spectral similarity. It has been used to segment wetland areas into distinct classes, such as water, marsh, swamp \cite{jing2020exploring}. However, K-Means assumes spherical clusters and requires the number of clusters to be predefined, which may not align with the complex structure of wetlands. Self-organizing maps (SOM) are neural network-based methods that project high-dimensional data onto a low-dimensional space, preserving topological relationships. They have been applied to wetland mapping to visualize spatial patterns and identify transitional zones \cite{kim2023classifying}. Despite their utility, SOMs are computationally intensive and may require expert interpretation. Unsupervised methods are advantageous for their ability to process unlabeled data, but they often lack the precision of supervised approaches and may produce less actionable results.  


\subsubsection{Deep learning}
In the field of wetland classification and mapping, deep learning methods have emerged as powerful tools for analyzing remote sensing data, offering significant potential for application. Current deep learning approaches commonly employed in wetland remote sensing include convolutional neural networks (CNNs), region proposal-based instance segmentation methods, fully convolutional networks (FCNs), Transformer-based models, and generative adversarial network (GAN)-based models. 

CNNs and their variants, such as ResNet \cite{he2016deep} and DenseNet \cite{huang2017densely}, extract spectral-spatial features from wetland remote sensing data through multiple layers of convolution and pooling. A CNN transforms an input remote sensing image \(X \in \mathbb{R}^{H \times W \times C}\) (with height \(H\), width \(W\), and \(C\) channels) into a set of high-level feature representations, ultimately producing a classification output. The core operation in a CNN is the convolution, which, for a given layer, can be expressed as:

\begin{equation}
Z_{i,j}^{(l)} = \sum_{u=0}^{k-1} \sum_{v=0}^{k-1} \sum_{c=0}^{C_{l-1}-1} W_{u,v,c}^{(l)} \cdot X_{i+u,\,j+v,\,c}^{(l-1)} + b^{(l)},
\label{eq:convolution}
\end{equation}

where \(X^{(l-1)}\) is the input to the \(l\)-th layer (with \(X^{(0)}\) being the original image), \(W^{(l)} \in \mathbb{R}^{k \times k \times C_{l-1}}\) is the convolution kernel (or filter) of size \(k \times k\), \(b^{(l)}\) is the bias term for the \(l\)-th layer, and \(Z^{(l)}\) is the linear output of the convolution at layer \(l\).

The ability of automatically learning hierarchical spatial features from raw image data through localized receptive fields, weight sharing, and deep feature extraction to capture spatial dependencies and patterns at multiple scales makes them particularly powerful for image-based tasks such as wetland classification. \cite{mahdianpari2018wetland} utilized deep CNN to extract wetlands from high-resolution images and outperformed RF both in efficiency and accuracy. \cite{yang2022identifying} employed DenseNet to classify urban wetlands. 

\textbf{Region proposal.} Region proposal-based instance segmentation frameworks, such as Faster R-CNN-based methods \cite{ren2015faster, yuan2024relational, yuan2023muren}, Mask R-CNN-based methods \cite{he2017mask} and Mask2Former-based methods \cite{cheng2022masked}, provide precise identification of wetland features, especially in ecosystems with complex and fragmented boundaries. The core of region proposal-based methods is the region proposal before feature extraction and segmentation: selective search algorithm extracts approximately thousands of class-agnostic region proposals $\mathcal{R} = \{R_j\}$ from an input image, where each region $R_j$ is parameterized by coordinates $P_j = (x_j, y_j, w_j, h_j)$, which can be simply considered as SNIC in a supervised manner, both focusing on the complete patches rather than pixels. By combining region generation with refined segmentation, these methods effectively address features such as marsh patches and water boundaries. \cite{yang2020applied} applies Mask R-CNN to segment water bodies to keep the integrality of the water. \cite{ma2022identifying} proposed cascaded R-CNN model to detect the dike-ponds with manual regular boundaries. 

\textbf{Fully convolutional network.} Fully convolutional networks (FCNs), with their encoder-decoder architecture, provide an end-to-end solution for pixel-level classification \cite{yuan2024fusu}. A more general formula is:
\begin{equation}
Y = \text{softmax} \left( f_{\text{decode}} \left( f_{\text{encode}}(X) \right) \right),
\label{eq:fcn}
\end{equation}
where \( f_{\text{encode}}(X) \) is the encoder, outputting multi-scale feature maps; \( f_{\text{decode}} \) is the decoder, generating a score map through upsampling and feature combination; \( \text{softmax} \) converts scores to probabilities.
These methods are typically trained end-to-end using a per-pixel cross-entropy loss:
\begin{equation}
\mathcal{L} = - \sum_{h=1}^H \sum_{w=1}^W \log Y_{h,w,\hat{Y}_{h,w}},
\label{eq:fcnloss}
\end{equation}
where \( Y_{h,w,k} \) is the predicted probability for pixel \( (h,w) \) in class \( k \), and \( \hat{Y}_{h,w} \) is the ground-truth label.

The principle of FCN involves extracting features with a fully convolutional network and generating a high-resolution segmentation map using upsampling and skip connections, laying the foundation for later models like U-Net \cite{ronneberger2015u, yuan2022melting} and PSPNet \cite{zhao2017pyramid, yuan2024fusu}, which have demonstrated excellent performance in wetland mapping, particularly in tasks that use high-resolution images such as WorldView and aerial images \cite{dang2020coastal, li2021mapping}. These methods effectively preserve pixel details and enhance the recognition of wetland structures through multi-scale perception. 

\textbf{Transformer.} In recent years, transformer-based deep learning models, such as Vision Transformer (ViT) and Swin Transformer, have gained attention for their ability in processing and understanding numerous remote sensing data. They leverage multi-head self-attention (MSA), which computes pairwise interactions between all patches. Given an input sequence \(\mathbf{Z} \in \mathbb{R}^{(N+1) \times D}\), self-attention is computed as:

\begin{equation}
\text{Attention}(Q, K, V) = \text{softmax} \left( \frac{Q K^T}{\sqrt{d_k}} \right) V,
\label{eq:self_attention}
\end{equation}

where \(Q=Z W_Q, \ K=Z W_K, \ V=Z W_V\) are the query (Q), key (K), and value (V) matrices; \(W_Q, Q_K, Q_V \in \mathbb{R}^{d \times d_k}\) are learnable weight matrices; \(d_k = d / h\) is the dimension per attention head for \(h\) attention heads. MSA makes them able to capture long-range dependencies between geographical features to be suitable for large-scale landscape analysis. \cite{jamali2022swin} combined Swin transformer and deep CNN for complex coastal wetland classification. \cite{marjani2025novel} proposed a Transformer-based method with spatio-temporal embedding to map wetlands in complex scenarios. Besides, these models are also adept at integrating multi-source data, improving robustness in wetland monitoring, particularly in cloud-affected regions \cite{jamali2021wetland, jamali20223dunetgsformer}. Additionally, transformer-based models show promise in few-shot learning, where pretraining and fine-tuning can mitigate the challenge of limited labeled wetland data \cite{jamali2022deep}. 

\textbf{Generative Adversarial Networks.}  
Generative Adversarial Networks (GANs) \cite{goodfellow2020generative} leverage the adversarial interplay between a Generator and a Discriminator to produce realistic data, and have emerged as a powerful tool for enhancing wetland mapping by addressing data scarcity and domain adaptation challenges. A GAN consists of two competing networks: a generator \( G \) and a discriminator \( D \), trained through a minimax game. The objective function is formulated as:  
\begin{equation}
\min_G \max_D \mathbb{E}_{x \sim p_{\text{data}}}[\log D(x)] + \mathbb{E}_{z \sim p_z}[\log(1 - D(G(z)))], \label{eq:gan_loss}  
\end{equation}  
where \( x \) represents real data samples from the data distribution \( p_{\text{data}} \); \( z \) represents random noise sampled from a prior distribution \( p_z \), and \( G(z) \) is the generated synthetic samples. \cite{jamali2021synergic} utilized GAN-based methods to generate synthetic wetland samples from the random noise vectors for Sentinel-1/2 images with a conditional map unit. \cite{he2019wetland} proposed a weak-supervised GAN-based method to classify wetlands. This generative manner provides researchers with a strong way to complete few-shot learning or weak-supervised learning. 



\subsection{Wetland mapping inventories}

\begin{table*}[t]
\caption{Summary of contemporary wetland inventories.}
\label{tab:product}
\renewcommand{\arraystretch}{1.5}
\resizebox{\linewidth}{!}{
\begin{tabular}{cccccc}
\hline
Scale                        &  Publication &Study Area & Category & Data Source & Mapping Year \\ \hline
\multirow{5}{*}{Regional}    & \cite{wang2024interannual}           &   Main urban areas, China       &     4 & Landsat, DEM       &       1985-2022 (yearly)       \\
                             &    \cite{han2015four}       &   Poyang Lake, China       &   4   &   Landsat, QuickBird    &    1973-2013 (yearly)         \\
                             & \cite{martins2020deep}           &    Millrace Flats Wildlife Management Area, USA      &  2  &    WorldView-3, LiDAR     &    2015          \\
                             &  \cite{baloloy2020development}     & Mangrove sites in Philippines and Japan    &   1       &    Sentinel-2         &   2016-2019           \\
                             &  \cite{louzada2020landscape}     &  Pantanal, Brazil   &     3     &    Landsat         &    1985-2019 (yearly)         \\ \hline
\multirow{5}{*}{National}    &    \cite{mao2020national}        &   China       &    14         &   Landsat    &   2014-2016 (yearly)    \\
                             &  \cite{gong2010china}          &  China        &  14  &    Landsat     &    1990, 2000          \\
                             &   \cite{niu2012mapping}         &  China        &   14&      Landsat     &    1978, 2008          \\
                             &   \cite{bwangoy2010wetland}  &    Congo   &     4     &     Landsat, JERS-1, DEM        &   1990-2000           \\
                             &   \cite{zhang2021fine}    &  China   &   1       &     Gaofen-1, Ziyuan-3        &     2018         \\ \hline
\multirow{5}{*}{Continental} &  \cite{tana2013wetlands}     &  North America        &    1         &  MODIS, DEM  &     2008     \\
                             &   \cite{wang2023wetland}         & East Asia     &  14   &    Sentinel-1/2, DEM         &     2020         \\
                             &  \cite{li2022mapping}   &   Africa   &    8      &     Landsat, DEM        &     2020         \\
                             &  \cite{liu2022precise}     &   Southeast Asia  &   11       &   Sentinel-2, DEM          &     2016-2024 (yearly)         \\
                             &  \cite{sun2024first}     &   South America  &    10      &    Sentinel-1/2         &     2020         \\ \hline
\multirow{5}{*}{Global}      &    \cite{pekel2016high}        & Global     &  1  &     Landsat        &     1984-2015 (monthly)         \\
                             & \cite{murray2019global}    &  Global     &    1      &     Landsat         &     1984-2016 (yearly)         \\
                             & \cite{lehner2024mapping}       &  Global  &    33      &     -        &    -          \\
                             &   \cite{zhang2024global}     &  Global  &    8      &    Landsat         &      2000-2022 (yearly)        \\
                             &   \cite{bunting2022global}    &   Global  &   1       &     Landsat        &     1996-2020 (yearly)         \\ \hline
\end{tabular}}
\end{table*}

Table \ref{tab:product} summarizes the successful applications of remote sensing technology in various wetland mapping applications in recent years. We divide them into four classes based on the study scale: regional, national, continental and global. 

At the regional level, the majority of studies concentrate on specific, typical, and crucial wetland areas, such as Poyang Lake in China \cite{han2015four} and the Pantanal wetlands in Brazil \cite{louzada2020landscape}. For example, \cite{han2015four} mapped four-decade changes of Poyang Lake in China using Landsat and QuickBird images yearly. \cite{martins2020deep} used high-resolution WorldView-3 and LiDAR images to map the wetlands in Millrace Flats Wildlife Management Area in the USA in 2015. Within this context, it is possible to identify categories of wetlands and to achieve classifications of internal land cover within these wetland areas. This efficiency primarily stems from the capacity to obtain high-resolution remote sensing images and ground samples at a reduced cost when conducting research in smaller experimental regions. Particularly, the recent advancements in drone technology have enabled the utilization of drones equipped with hyperspectral sensors and LiDAR sensors to capture imagery characterized by both high spatial and high spectral resolution \cite{adam2010multispectral}. Such images encompass rich spatial and spectral information, facilitating comprehensive internal land cover classification within wetland environments.

At the national scale, studies often focus on producing comprehensive wetland inventories to support national environmental governance and ecological conservation policies. For instance, \cite{gong2010china} developed China's first multi-year 30-m resolution national wetland map in 1990 and 2000, classifying 14 wetland categories using Landsat time-series data. Similarly, \cite{mao2020national} and \cite{niu2012mapping} established baseline wetland databases for China in 2014/2015/2016 and 1978/2008 respectively, both employing Landsat imagery to systematically monitor wetland changes across this vast territory. These national-scale mappings typically employ medium-resolution satellite data (e.g., Landsat, Sentinel) to balance spatial detail and computational feasibility, occasionally supplemented by high-resolution imagery for critical regions \cite{zhang2021fine}. For example, \cite{zhang2021fine} integrated Gaofen-1 and Ziyuan-3 satellite data to map mangroves in China with enhanced spatial precision. 
The emergence of cloud computing platforms like Google Earth Engine has significantly enhanced the processing capabilities for national-scale analyses, enabling efficient handling of massive satellite data volumes and multi-temporal analyses \cite{mao2020national}. 

At the continental scale, wetland mapping efforts typically aim to harmonize multi-country datasets while addressing diverse bioclimatic conditions and regional classification challenges. For example, \cite{tana2013wetlands} mapped North American wetlands using MODIS and DEM data, focusing on broad-scale inundation patterns, while \cite{li2022mapping} classified eight wetland types across Africa by integrating Landsat imagery and topographic data. Recent advances have leveraged multi-sensor fusion to improve accuracy: \cite{wang2023wetland} combined Sentinel-1 SAR and Sentinel-2 optical data with DEM derivatives to map 14 wetland categories across East Asia. Challenges include reconciling varying national wetland definitions, mitigating seasonal data gaps in monsoon regions, and managing computational demands for processing petabyte-scale satellite archives. However, validation remains problematic due to inconsistent ground truth availability across countries, particularly in transboundary wetland complexes like the Congo Basin \cite{bwangoy2010wetland}.  

At the global level, wetland mapping initiatives face unprecedented challenges in balancing spatial detail, consistency across ecosystems and unified classification system. We can notice that most inventories in this level focus on one or some specific categories, hard to achieve a fine-grained classification system. For example, \cite{pekel2016high} generated monthly 30-m water extent maps globally from 1984 to 2015, while \cite{murray2019global} focused on global annual tide flat changes. \cite{zhang2024global} achieved annual 30-m global wetland maps from 2000 to 2022 by Landsat data with machine learning methods and yearly training samples. One exception is \cite{lehner2024mapping}, which combined all other wetland inventories to map the global wetlands in a 33-category classification system. 


\subsection{Scientific importance}\label{sec:support}
Wetland mapping products have emerged as critical tools in environmental science, providing high-resolution spatial data on the distribution, type, and extent of wetland ecosystems. These products are particularly instrumental in supporting scientific research, including methane emission estimation, biodiversity conservation, water quality management, flood control, etc. Leveraging data from remote sensing data and other auxiliary data, these products offer precise, scalable insights that enhance our understanding of wetland functions and their role in global environmental systems.

\subsubsection{Methane emission estimation}
Wetland mapping products are crucial for estimating methane emissions, a significant greenhouse gas contributing to climate change. Wetlands, especially those with waterlogged soils, are major natural sources of methane due to anaerobic decomposition \cite{zhang2017emerging}. Wetland inventories provide critical data on wetland distribution and classification, which are used in models to estimate methane production and emission rates. For instance, \cite{xiao2024global} used the IBIS-CH4 model, integrating wetland mapping data to simulate global emissions from 2001 to 2020, averaging $152.67 \ Tg \ CH_4 \ yr^{-1}$. This highlights their role in identifying potential emission sources and quantifying wetland extent for emission calculations. Another study \cite{zhang2017emerging} discussed the role of wetland methane in climate change. By an ensemble estimate of wetland $CH_4$ emissions driven by 38 general circulation models for the 21st century, it suggests that climate mitigation policies must consider mitigation of wetland $CH_4$ feedbacks. 

The importance of spatial resolution in modeling methane emissions was emphasized by \cite{albuhaisi2023importance}, which investigated the impact of high-resolution wetland maps on emission estimates for the Fennoscandinavian Peninsula. Additionally, a study using Landsat ETM+ data for modeling methane emissions in Australia \cite{akumu2010modeling} demonstrated the utility of mapping in local-scale emission estimates. These examples illustrate how wetland mapping supports methane studies by providing detailed spatial and temporal wetland distribution and variation data, crucial for understanding emission dynamics and informing policy.

\subsubsection{Biodiversity conservation}

Wetlands are among the most biodiverse ecosystems, providing habitats for numerous plant and animal species, including many that are endangered or migratory. Wetland mapping plays a critical role in biodiversity conservation, serving as a fundamental tool for identifying, monitoring, and protecting ecologically significant wetland areas. They host an estimated 40\% of global biodiversity, encompassing plants, insects, amphibians, and birds. By accurately defining wetland boundaries and types, mapping enables conservationists and policymakers to identify biodiversity hotspots—regions with exceptionally high species diversity—where targeted protection and management efforts can be prioritized.


\cite{yi2024global} proposed a cost-effective assessment model using mapping data to identify conservation priorities, covering 28\% of global wetland distribution, with only 44\% currently protected. \cite{ma2010managing} discussed how wetlands can be managed to provide waterbird habitats. These habitat variables include water depth, water level fluctuation, vegetation, salinity, topography, food type, food accessibility, wetland size, and wetland connectivity, which needs detailed wetland mapping data and field surveys. \cite{dertien2020relationship} examined the relationship between proportional wetland cover and species richness across the conterminous United States, addressing the uncertainty surrounding this association at the sub-continental scale. Using data from the National Wetlands Inventory (NWI) and the National Land Cover Database (NLCD), wetland cover and its changes between 2001 and 2011 were analyzed. A Bayesian spatial Poisson model was applied to estimate the spatially varying effect of wetland cover on the richness of amphibians, birds, mammals, reptiles, and terrestrial endemic species. Furthermore, \cite{ehrenfeld2008exotic} discussed wetland mapping in invasive species management within biodiversity conservation, which used mapping to track invasive species affecting wetland biodiversity, often overlooked in broader conservation strategies.

\subsubsection{Water quality assessment}
Wetlands play a significant role in water quality assessment by filtering pollutants and improving water clarity. Mapping products help assess their contribution by providing data on wetland location and type, which influence purification processes. \cite{ferreira2023wetlands} highlighted their potential to remove organic and inorganic pollutants through physical, chemical, and biological processes. \cite{kadlec2008treatment} used mapping to evaluate the effectiveness of constructed wetlands in pollutant removal, showing higher efficiency in areas with dense vegetation cover. \cite{gholizadeh2016comprehensive} discussed the utility of remote sensing data, including wetland mapping products in water quality assessment, which not only requires the large-scale data, but also high-accuracy local-scale data. Another study \cite{land2016effective} used wetland mapping data to compare nutrient removal rates across different wetland types, emphasizing the role of mapping in management planning.

\subsubsection{Flood control}
Wetland mapping is crucial for flood control and hydrological studies, identifying areas that store floodwater and reduce downstream impacts. Wetlands act as natural sponges, absorbing excess water during storms and releasing it slowly, mitigating flood damage. \cite{wu2023wetland} used hydrological modeling with wetland modules to quantify their regulation efficiency, showing significant flood attenuation in specific areas. \cite{pattison2018wetlands} also emphasizes the role of wetlands in storing runoff, particularly in upper watersheds, reducing flood-related problems in Canada.

Mapping products enable planners to locate these critical areas, enhancing flood risk management and hydrological connectivity analysis. \cite{acreman2007hydrological} used mapping to assess wetland restoration impacts on flood control, showing improved water retention. \cite{erwin2009wetlands} highlighted mapping's role in understanding hydrological changes under climate scenarios, often linking flood control to broader climate impacts.

\subsection{Driving factors}\label{sec:factor}
Wetlands are ecologically vital ecosystems, providing essential services such as biodiversity conservation, flood regulation, and water purification. However, they are undergoing significant global decline, driven by a complex interplay of natural and anthropogenic factors. Understanding the key drivers behind wetland loss is crucial for effective conservation planning. This section synthesizes wetland classification and mapping-related publications to examine the primary forces contributing to wetland degradation and retreat. 
\subsubsection{Anthropic land use change}
Wetland destruction worldwide has been primarily driven by anthropic land use changes such as land reclamation and drainage for agricultural purposes, the construction of infrastructure to support urbanization, and the expansion of planted forests for commercial or conservation goals, etc. Significant wetland losses have been documented in regions such as Southern Ontario (80–90\% destroyed), Europe (60–90\% lost in the last century) \cite{penfound2022analysis,verhoeven2014wetlands}, and semi-arid South Africa (over half destroyed by agriculture, mining, and urban development) \cite{gxokwe2020multispectral}. In tropical and subtropical countries like Brazil and Indonesia, export-oriented agro-industries have led to the destruction of mangroves (up to 75\% lost) and peat swamps (up to 50\% converted to oil palm plantations) \cite{asselen2013drivers, NWI2019}.
For instance, the U.S. Fish and Wildlife Service's National Wetlands Inventory (NWI) report highlights that between 2009 and 2019, wetland loss in the United States was primarily associated with urban development, the establishment of upland planted forests, and agricultural expansion \cite{NWI2019}. \cite{han2015four} examined the land use changes in 40 years of Poyang Lake, to identify the effects of historical human activities on the changes of Poyang Lake's morphology and extent. This trend extends a long-term pattern of net wetland loss observed over previous decades, underscoring the persistent pressure exerted by human activities on these critical ecosystems.

\subsubsection{Construction of dams and reservoir} The construction of reservoirs and dams poses a significant threat to wetlands globally, as they alter river discharge patterns, sediment loads, and ecological conditions in floodplains \cite{junk2013current}. With over 38,000 large reservoirs worldwide \cite{mulligan2020goodd}, their primary purposes include irrigation, hydropower generation, flood control, and water supply. Hydropower is a major electricity source in many regions, such as South America (e.g., Brazil, Paraguay, Peru) and Africa, where it contributes over 50\% of electricity in 25 countries \cite{del2014latin}. Similarly, in the Pacific, countries like Fiji and New Zealand rely heavily on hydropower. However, these structures disrupt longitudinal and lateral connectivity between rivers and their floodplains, leading to severe ecological impacts, including biodiversity loss and the degradation of floodplain functions. In Europe and North America, most large rivers are accompanied by dikes, resulting in the destruction or heavy modification of floodplains. \cite{acreman2007hydrological} highlighted how water management practices, like dams, can lead to wetland desiccation. In South America, reservoirs on rivers like the Magdalena and upper Paraná have degraded floodplains, and similar trends are expected in Africa, China, and tropical Asia. Additionally, rising water demand in the Okavango River's upper catchment threatens the iconic Okavango Delta \cite{andersson2003water, wantzen2024end,edmonds2020coastal}. Overall, the continued expansion of reservoirs and flood defenses poses a growing risk to wetland ecosystems worldwide. \cite{han2015four} also reported that in recent years, large-scale hydrological modifications, particularly the impoundment of the Three Gorges Dam, have further exacerbated changes in downstream wetland morphology and extent, leading to shifts in ecological dynamics and biodiversity. These examples illustrate the complex interplay between direct human actions and indirect environmental changes in driving wetland conversion, highlighting the need for integrated conservation strategies to mitigate further loss and degradation.

\subsubsection{Climate change} Global climate change, manifested through altered precipitation patterns, rising temperatures, and sea level rise, affects wetland hydrology and ecology. \cite{junk2013current} noted that climate change disrupts wetland functions, particularly in regions experiencing increased drought or flooding. Climate change impacts on wetlands vary by region, with higher latitudes experiencing greater temperature rises that will significantly affect biota, potentially leading to the forestation of boreal wetlands. In contrast, tropical and subtropical regions will face more pronounced effects from changes in precipitation patterns and extreme climate events, such as increased cyclones, floods, droughts, and fires, driven by phenomena like El Niño, La Niña, and monsoon variability. However, predictions remain uncertain due to imprecise global climate models and insufficient regional data. While tropical and subtropical wetland species may adapt due to their wide distribution ranges and physiological resilience, coastal wetlands face severe threats from rising sea levels, which could squeeze them between advancing seas and human infrastructure. Coastal wetlands are particularly critical due to their high biodiversity, ecological connectivity between land and ocean, and the essential services they provide to humans.

\begin{figure*}[t]
    \centering
    \includegraphics[width=1.0\linewidth]{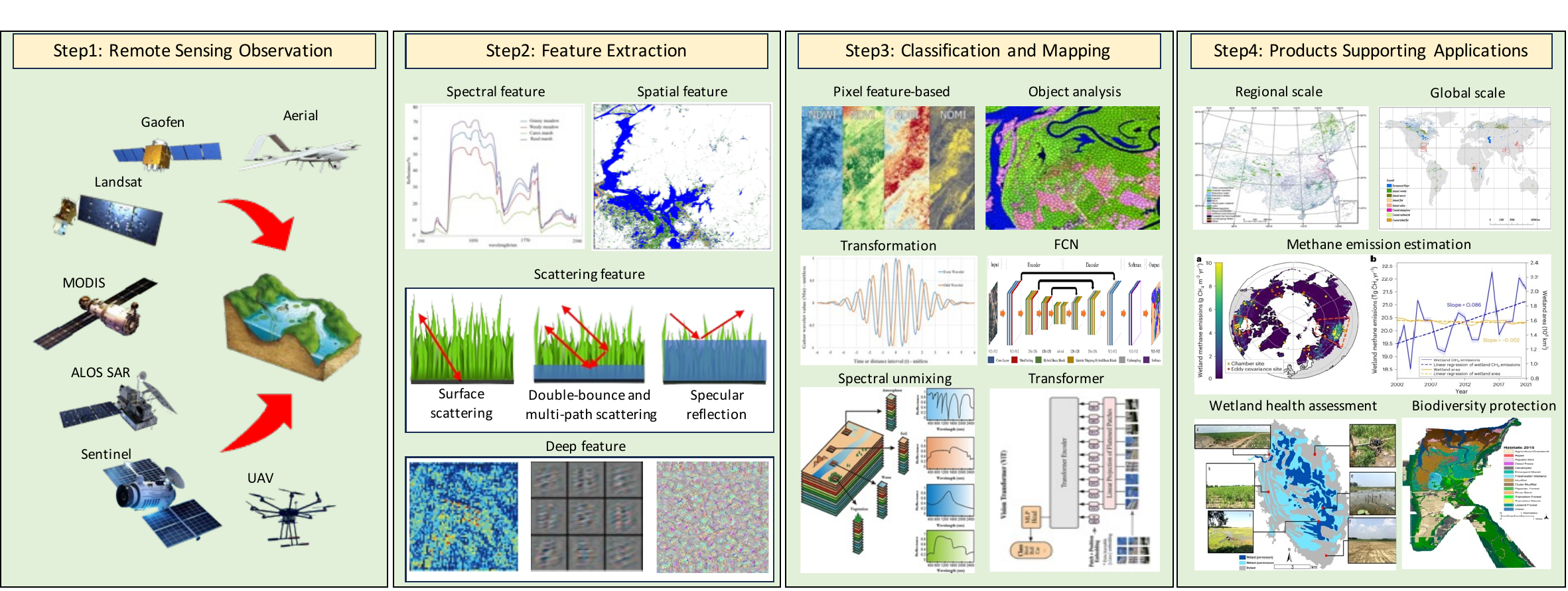}
    \caption{The current research paradigm of wetland classification and mapping, including remote sensing observation, feature extraction, classification and mapping, products supporting applications. Pictures are from \cite{ballanti2017remote}, \cite{singh2021hydrogeomorphic}, \cite{yuan2024boreal}, \cite{gong2010china}, \cite{zhang2024global}, \cite{dosovitskiy2020image}, \cite{zhang2022spatiotemporal}.}
    \label{fig:current}
    \vspace{0em}
\end{figure*}

\subsubsection{Ecological invasion} wetlands are increasingly threatened by the introduction of exotic plant and animal species, often due to negligence, deliberate release, or global trade. Many invasive species, such as the Brazilian pepper, Australian paper bark tree, and Asian tiger python, have caused severe environmental problems in places like the Everglades \cite{ren2021invasive, rodgers2014mapping}. Additionally, species like the Asian golden mussel, introduced via ballast water, have rapidly spread and disrupted local aquatic systems. Climate change is expected to exacerbate these issues by stressing native species and creating conditions that favor invasive species, including disease vectors, further threatening wetland biodiversity, leading to wetland restoration and decline \cite{erwin2009wetlands}.

\section{Discussion}\label{sec:discussion}
\subsection{Current research paradigm} \label{sec:current}

Contemporary wetland mapping approaches increasingly rely on hybrid methodologies and multi-source data fusion to address the inherent complexity of wetland ecosystems (Fig. \ref{fig:current}). Single-algorithm frameworks (e.g., pure deep learning or pixel-based classifiers) often struggle to capture the spectral, structural, and hydrological heterogeneity of wetlands, particularly when distinguishing between spectrally similar classes like marshes, swamps, and seasonally inundated grasslands. To mitigate these challenges, researchers are integrating complementary techniques—for instance, combining machine learning classifiers (e.g., Random Forests), embedding feature engineering (e.g., texture indices, phenological metrics) and hierarchical decision trees into the overall workflows to form a multi-stage classification method \cite{wang2023wetland, liu2022precise}. This hybrid paradigm capitalizes on the robustness of machine learning classifiers and the interpretability of decision trees with manual configurations. However, such method integration often entails extensive manual intervention, which can introduce subjectivity and inefficiencies. A significant limitation lies in the sequential dependency of subsequent processing steps on the initial classification stage. If the feature representation in the first stage lacks sufficient discriminative power or fails to capture the full spectrum of wetland characteristics, the overall performance of the workflow may be compromised. This underscores the need for more automated and adaptive frameworks that enhance feature representation while minimizing manual adjustments, thereby improving the reliability and scalability of wetland classification and mapping workflows.

Similarly, data fusion has become indispensable. Optical imagery (e.g., Sentinel-2, Landsat) provides rich spectral information but is limited by cloud cover and temporal gaps, whereas SAR (e.g., Sentinel-1, NISAR) offers all-weather, day-night capabilities to track hydrological dynamics. Fusion of optical, SAR, and ancillary data (e.g., LiDAR-derived topography, soil moisture maps) enables holistic characterization of wetland properties, such as inundation patterns, vegetation structure, and subsurface hydrology. For example, SAR coherence can detect subtle water-level changes beneath dense canopies, while hyperspectral data resolves biochemical indicators critical for water quality assessments.

\subsection{Critical gaps between methods and data}\label{sec:gap}
Despite advances, critical gaps undermine the operationalization of wetland mapping for dynamic environments. \textbf{First}, while multi-source data availability has surged, most algorithms inadequately exploit cross-sensor synergies. Due to differences in imaging mechanisms and physical representations—such as optical multispectral remote sensing images being acquired through passive sensors, while DEM relies on active polarimetric SAR technology—the significant domain gaps between multi-source heterogeneous remote sensing data limit the effectiveness of data fusion \cite{schmitt2016data}. To address this, existing research has explored various fusion techniques based on multi-source remote sensing imagery. Most studies directly employ feature stacking-based methods for modality fusion \cite{wang2023wetland}. However, this approach increases the feature dimensionality and computational complexity of the samples. In contrast, subspace-based methods address this issue by representing heterogeneous data in low-dimensional subspaces, thereby avoiding the curse of dimensionality in subsequent classification tasks. Many subspace-based methods utilize techniques such as IHS transformation and PCA. These methods effectively reduce feature dimensionality, decrease computational load, and improve the signal-to-noise ratio of fusion. Nevertheless, their generalizability in real-world scenarios and their ability to extract nonlinear relationships between different data sources remain limited. Therefore, it is necessary to develop multi-source heterogeneous remote sensing data fusion methods capable of effectively modeling cross-modal relationships and decomposing both shared and unique features across modalities. These methods should also demonstrate broad applicability and effectiveness in wetland classification and extraction tasks. \textbf{Second}, the lack of dynamic training samples severely limits temporal mapping fidelity. Current workflows rely on static, often outdated ground truth, failing to capture dynamic wetland transitions (e.g., hydroperiod shifts, vegetation phenology). Even some research, for example, \cite{zhang2024global} utilized single-temporal global wetland mapping products and three-temporal global land cover products to generate potential samples. They then employed a change detection model to remove samples exhibiting changes, resulting in the final multi-temporal samples. However, the number of samples generated by such methods gradually diminishes over time, and the direct removal of samples introduces temporal discontinuities, which are detrimental to model training and the continuity of mapping products. Consequently, there remains significant room for improvement in the flexibility and effectiveness of current automated label generation methods. Emerging solutions include self-supervised learning (SSL) techniques, such as contrastive predictive coding, which leverage unlabeled time-series data to learn invariant wetland features across seasons, or generative adversarial networks (GANs) to synthesize realistic temporal training patches.
\textbf{Third}, temporal inconsistency plagues existing time-series products \cite{wang2015mapping}. Many approaches process images independently, ignoring phenological trajectories or antecedent hydrological conditions. For example, a marsh classified as permanently-inundated in June may transition to saturated soil by August, but disjointed classifications fail to encode these trajectories, undermining applications like methane flux modeling. 
\textbf{Fourth}, the scarcity of large-scale, fine-grained pixel-wise annotations remains a bottleneck. Public datasets (e.g., NWI) suffer from coarse class definitions, geographic biases and coarse annotation (e.g., point-based rather than pixel-wise polygon-based). 

\subsection{Limitations of current wetland mapping products}\label{sec:limitation}

\begin{figure}[h]
    \centering
    \includegraphics[width=1.0\linewidth]{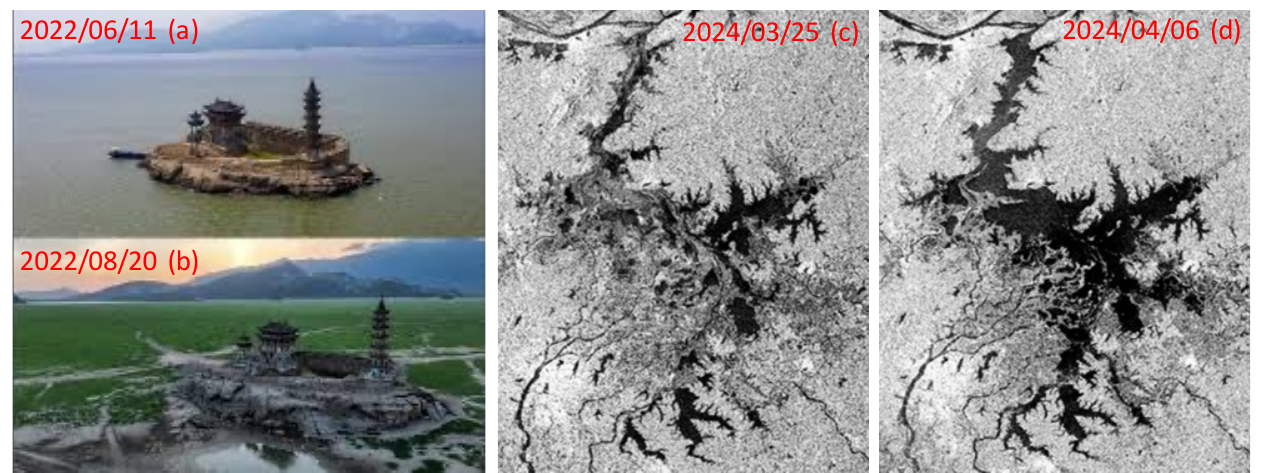}
    \vspace{-1.5em}
    \caption{The high dynamics of Poyang Lake, China. (a) and (b) are captured by UAV. (c) and (d) are captured by Sentinel-1.}
    \label{fig:dynamic}
    \vspace{0em}
\end{figure}

Existing wetland maps, while valuable, exhibit three systemic shortcomings that curtail their utility in addressing global environmental challenges. \textbf{First}, the near-universal absence of dynamic wetland products obscures critical hydrological and ecological processes. Static annual composites cannot resolve phenological inundation pulses (e.g., Poyang Lake in Fig. \ref{fig:dynamic}), ephemeral wetland formation after extreme rainfall, or drought-induced fragmentation—all of which are pivotal for quantifying greenhouse gas budgets or assessing habitat connectivity for migratory species. \textbf{Second}, a persistent gap exists between spatial scale and classification specificity. Global initiatives like the Global Lakes and Wetlands Database (GLWD) \cite{lehner2004development} prioritize broad coverage but have limited categories of wetlands, obscuring functional diversity (e.g., tidal vs. floodplain). Conversely, regional high-resolution maps capture fine-scale heterogeneity but lack compatibility with Earth system models requiring globally consistent inputs. A hierarchical classification system—featuring Level 1 classes (e.g., swamp) for global coherence and Level 2-3 subclasses (e.g., tidal mangrove) for regional specificity—could reconcile this tension. \textbf{Third}, current products remain siloed from downstream applications, limiting their operational value. For instance, for flood mitigation, static wetland extents cannot inform real-time floodwater storage capacity during storms, necessitating dynamic inundation layers fused with hydraulic models.

\section{Prospects}\label{sec:pros}
Based on the aforementioned meta-analysis, review and in-depth discussion, wetland classification and mapping prospects have emerged from this attempt, which concerns past, current and future.

\subsection{Multi-source and multi-resolution remote sensing data fusion}
Over the past six decades, remote sensing has evolved into an indispensable tool for characterizing wetlands, which are inherently complex, spatially heterogeneous, and temporally dynamic ecosystems \cite{guo2017review}. Today, there is a growing consensus that no single data source can fully capture the multifaceted nature of wetlands, especially when considering variations in hydrological connectivity, vegetation assemblages, and geomorphology across scales ranging from small marsh patches to entire watersheds. As a result, fusing multi-source and multi-resolution remote sensing data has become a core strategy in wetland research, aiming to balance the strengths and limitations of each sensor and resolution level while delivering comprehensive insights into wetland dynamics \cite{li2022deep}.

From the perspective of multi-resolution imagery, high spatial resolution data are crucial for detailed ecological assessments, such as delineating subtle transitions at wetland boundaries, detecting microtopographical features in peatlands, or monitoring narrow tidal creeks in coastal marshes. Unmanned aerial vehicles (UAVs) and high-resolution satellite sensors can capture fine-scale patterns, enabling researchers to investigate localized phenomena, such as the distribution of rare wetland plant species. However, high-resolution data typically covers limited extents and may be prohibitively expensive or difficult to acquire consistently over large regions or extended time periods. Medium- to low-resolution satellite imagery, while providing broader coverage and more frequent revisit times, often lacks the granularity needed to characterize small or fragmented wetlands. Consequently, combining high-resolution data for precise local mapping with moderate-resolution data for larger-scale ecological context offers a more holistic depiction of wetland systems. 

In terms of multi-source integration, optical imagery remains a cornerstone in wetland monitoring for its ability to capture spectral signatures associated with vegetation vigor, water quality, and sediment concentration. Yet frequent cloud cover, tidal fluctuations, and the prevalence of inundated surfaces can limit optical data’s usability. SAR images penetrate cloud and darkness, furnishing information on wetland inundation status, emergent vegetation structure, and surface roughness, even under challenging atmospheric conditions. LiDAR contributes detailed elevation profiles and canopy height measurements, elucidating subtle topographic gradients critical to wetland hydrology and plant distribution. Hyperspectral sensors offer finely resolved spectral bands to discriminate subtle species- or substrate-level variations. By harmonizing data from these disparate sources, researchers can build robust, multi-dimensional representations of wetland environments that capture spatial, temporal, and biophysical variability. To achieve this, researchers must develop specific methods to bridge the domain gaps between multi-source heterogeneous remote sensing data, which are due to differences in imaging mechanisms and physical representations. 

Looking ahead, emerging technologies and platforms promise to further expand the potential of multi-source, multi-resolution data fusion in this field. Advances in sensor miniaturization, computing capacity, and artificial intelligence algorithms will facilitate near-real-time data assimilation and automated feature extraction. Ground-based sensor networks and citizen science initiatives can also provide supplementary in situ measurements, enhancing the ecological context of remotely sensed observations. Ultimately, a synergistic approach that integrates diverse data sources—ranging from broad-scale satellite constellations to finely resolved UAV flights—will drive a deeper understanding of wetland processes, improve mapping accuracy, and support more informed conservation and management strategies.

\subsection{Knowledge-guided explainable deep learning method for wetland classification and mapping}

Wetlands encompass a rich variety of types, exhibit diverse morphologies, and often possess indistinct boundaries, necessitating expert knowledge-based interpretation for accurate mapping and assessment. Over the past few decades, numerous researchers have sought to develop automated remote sensing interpretation algorithms to fully or partially replace traditional manual interpretation through artificial intelligence. Research into AI-driven techniques for automated remote sensing interpretation has undergone several waves of innovation, though early approaches such as random forest tended to rely on handcrafted features and models, which primarily capture basic morphological, textural, or spectral attributes \cite{wang2023wetland, zhang2024global}. Consequently, such methods were limited in representing more abstract semantic characteristics and often struggled with the complexity and diversity inherent to wetlands.

More recently, deep learning based on the evolution of computation ability and scaling ability has achieved transformative breakthroughs. These methods leverage large-scale training datasets to automatically learn and extract rich semantic representations from input data \cite{jamali20223dunetgsformer, hosseiny2021wetnet}. While some progress has been made in applying deep learning to wetland remote sensing, existing studies frequently adopt frameworks inherited from computer vision applications and face constraints in two aspects. First, due to the scarcity of dedicated, large-scale wetland remote sensing labeled training datasets, deep learning-based wetland classification and mapping methods can only perform well in limited scenarios. Second, as we discussed before, purely deep learning methods, although effective at processing vast, multi-source remote sensing imagery, often lack interpretability and do not fully exploit domain knowledge specific to wetlands, such as ecological, hydrological, or geomorphological insights. Different from other geographical objects like roads, buildings, forests, the complicated horizontal and vertical structures, the complex temporal features of wetlands make it difficult to delineate and classify wetlands without any domain knowledge. Therefore, to achieve the better mapping efficiency and results for wetlands, we need to develop a knowledge-guided paradigm, complementing the data-driven approach by integrating domain knowledge graphs and other relevant geoscientific information. This strategy enhances model interpretability and enables more meaningful inference. By combining remote sensing features, expert heuristics, and wetland-specific knowledge with AI methodologies, researchers can advance wetland remote sensing toward intelligent monitoring systems that blend both data-driven and knowledge-guided paradigms.

\subsection{Foundation model for wetland classification and mapping}
Building on the paradigm shift toward knowledge-guided artificial intelligence, there is a growing impetus to develop foundation models that can serve as broadly applicable baselines for various remote sensing tasks. In the context of computer vision and natural language processing, foundation models, large-scale neural networks pre-trained on diverse and extensive datasets, have proven effective at capturing robust, transferable representations. Analogously, a foundation model for remote sensing would involve training a universal deep learning architecture on vast amounts of remote sensing data, encompassing different sensors, phases and resolutions, to learn generalized patterns of geographical ecosystems \cite{guo2024skysense,zhang20242}. After post-training on the specific spectral, morphological, temporal patterns of wetlands, which would include the nuanced spectral profiles of emergent vegetation, the temporal dynamics of seasonal inundation, and the subtle textural cues indicative of wetland degradation or succession, we can expect the foundation models for wetland classification and mapping have emerging and special understanding of wetland patterns with a good generalization ability for all regions in the world. 

A critical advantage of deploying a foundation model lies in its transferability. Once pretrained on extensive remote sensing and wetland-related data, the model could be fine-tuned for specific downstream applications such as delineating wetland boundaries, classifying wetland types, monitoring invasive species, or assessing restoration progress. This approach reduces the burden of curating large, labeled datasets for each individual task, as the foundational knowledge embedded within the model can be adapted to new scenarios with substantially fewer labeled examples. Moreover, the model’s broad exposure to multi-resolution and multi-source imagery, ranging from high-altitude, low-resolution satellites to fine-grained UAV observations, can capture both local-scale heterogeneity and regional connectivity.

From a knowledge-guided AI perspective, foundation models can be further enriched through explicit integration of domain knowledge—such as ecological, hydrological, and geomorphological rules—into their training or inference processes. Rather than relying exclusively on “data-driven” pattern recognition, the model could incorporate constraints derived from wetland succession theory, topographic gradients, or known species habitat requirements, enhancing both interpretability and reliability. For instance, attention modules or graph-based encoding layers could be designed to reflect geospatial relationships among wetlands, facilitating informed reasoning about how local changes propagate through larger hydrological or ecological networks.

In practical terms, realizing a wetland-focused foundation model poses four challenges. First, assembling a sufficiently large and diverse training corpus requires coordination among international satellite missions, local agencies, and open-data consortia. The "Scaling Law" is losing its power on the pertaining stage \cite{guo2025deepseek}. Therefore, high-quality, well-structured, well-curated data is needed. Second, the model’s architecture must accommodate distinct data modalities (e.g., hyperspectral bands for vegetation discrimination and SAR backscatter for inundation mapping) while also being computationally efficient enough to handle multi-temporal analyses. Third, back to the advances in large language models, the emerging intelligence is found to come over during the reinforcement learning (RL) period after the supervised fine-tuning stage \cite{guo2025deepseek}. To gain a strong and thoughtful large model for wetland classification and mapping, we must consider how to construct the specific RL stage for wetland knowledge. Fourth, robust evaluation protocols are needed to ensure that the model’s performance generalizes across a variety of wetland types and environmental conditions.

Nevertheless, the promise of a foundation model approach to wetland monitoring is substantial. By synthesizing multi-source, multi-temporal, multi-resolution remote sensing data with domain-specific knowledge, such models can accelerate breakthroughs in wetland science, enabling near-real-time ecological monitoring, refined estimates of wetland ecosystem services, and proactive management strategies aimed at conservation and sustainable use. Over time, these advances may also foster greater interdisciplinary collaboration, as datasets from hydrology, climatology, and socioeconomics are integrated into a unified AI-driven framework for understanding and preserving Earth’s vital wetland ecosystems.

\subsection{Task-oriented adaptive mapping for diverse downstream applications}  
Future wetland mapping must evolve beyond generic, one-size-fits-all approaches to deliver tailored solutions that address the distinct requirements of specific downstream applications. Wetlands play multifaceted roles in methane emission regulation, water quality maintenance, flood mitigation, biodiversity conservation, and carbon sequestration, each demanding unique mapping parameters in terms of spatial resolution, classification granularity, temporal coverage, and update frequency.

A flexible classification system is critical to align mapping outputs with application-specific goals. While broad categories suffice for biodiversity assessments, methane emission studies demand finer distinctions, such as separating Typha-dominated marshes (high methane producers) from forested swamps (lower emissions). Similarly, water quality monitoring necessitates spectral and biophysical indicators integrated into wetland maps, whereas carbon stock assessments require stratification by vegetation biomass and soil organic content. Developing modular classification frameworks where core wetland types are supplemented by adjustable ancillary layers will enable seamless adaptation to diverse end-user needs.  

Spatial resolution is a critical factor that must be strategically aligned with application-specific objectives. For methane emission modeling, moderate spatial resolution (10–30 m) is often sufficient to identify methane hotspots such as waterlogged peatlands or anaerobic soils while balancing computational efficiency for large-scale analyses. In contrast, flood control applications require high-resolution mapping (<5 m) to delineate fine-scale hydrological features, such as narrow drainage channels, levees, or micro-topographic variations that influence water flow during extreme events. Biodiversity conservation efforts may prioritize intermediate resolutions (5–10 m) to map habitat patches and ecological corridors while maintaining regional coverage. Emerging technologies, such as fusion of data with different resolutions, enable flexible scaling of spatial resolution to match these needs. For example, blending 3 m PlanetScope imagery with 10 m Sentinel-2 bands can enhance feature discrimination in heterogeneous wetland mosaics, while AI-based super-resolution techniques may upscale coarse data for localized studies.

Temporal resolution and spatial resolution design should also be application-driven. Long-term time series are essential for quantifying wetland loss trends and climate feedbacks, whereas short-term, high-frequency observations are vital for disaster response. High spatial resolution data is important to detect fine-scale hydrological features, such as narrow drainage channels, levees, or micro-topographic variations that influence water flow during extreme events; while medium-resolution data is more suitable for large-scale mapping. For instance, methane emission modeling requires high temporal frequency (e.g., daily to weekly) to capture seasonal hydrological shifts and vegetation dynamics that drive gas fluxes, coupled with medium-to-high spatial resolution (<30 m) to identify methane hotspots like saturated soils or anaerobic peatlands. In contrast, flood control applications prioritize near-real-time monitoring (sub-daily updates) at high spatial resolution (<10 m) to track water level changes, inundation extent, and connectivity to river systems during extreme weather events. 

We should note that higher temporal and spatial resolution generally leads to better analytical capabilities, but practical applications often require trade-offs. In the "impossible triangle" of high temporal frequency (daily), high spatial resolution (<10m), and large-scale coverage (global), under the circumstance where the data processing ability of both hardware and software has limitations, it is extremely challenging to achieve all three simultaneously. Therefore, we should make strategic trade-offs based on the needs of downstream tasks to optimize results as much as possible.

We maintain a firm conviction regarding the future trajectory of technological advancements that facilitate seamless access to multi-source remote sensing data, cloud computing, edge processing, and reliable, explainable artificial intelligence, all while minimizing inference costs. These innovations are anticipated to support a task-oriented paradigm and potentially overcome the limitations of the so-called impossible triangle. Furthermore, models tailored to specific applications, such as those focusing on methane flux-correlated spectral traits or indicators of flood risk, have the capacity to automate the creation of customized wetland products. Nevertheless, the realization of scalability necessitates robust global collaboration to establish standardized, application-driven metadata protocols, such as the classification of water permanence types for hydrologic models, and to enable the sharing of curated training datasets across various disciplines. By emphasizing adaptive, purpose-driven mapping strategies, the remote sensing community possesses the opportunity to evolve wetlands from static cartographic representations into dynamic, decision-ready geospatial resources that significantly enhance climate action, ecosystem service valuation, and policy implementation.

\subsection{Spatiotemporal expansion of wetland mapping}
Historically, wetland remote sensing efforts have largely centered on regional or national scales, with only a limited number of studies addressing continental or global-scale mapping. Even high-profile initiatives, such as the NWI in the U.S., often lack sufficient update frequencies to capture wetland ecosystems’ intrinsically dynamic nature. Given that wetlands are driven by seasonal hydrological fluctuations, climate variability, and anthropogenic pressures, low-frequency updates fail to represent the rapid and sometimes unpredictable changes in wetland extent and condition. Consequently, developing high-frequency, large-scale wetland maps is critical for improving wetland conservation, management, and policy interventions.

To achieve consistent and robust global wetland mapping, a unified classification system is indispensable. Currently, discrepancies in classification schemes between countries, agencies, and research groups hinder cross-regional comparisons and data integration efforts. Establishing a standardized framework that aligns nomenclature, hierarchical structure, and ecological definitions would ensure more accurate and comparable inventories of wetland distribution and types worldwide. Such an approach would not only facilitate large-scale wetland monitoring but also ensure that regional or national wetland products can be effectively synthesized into a coherent global dataset.

Moreover, the spatial and temporal dimensions of wetland mapping require simultaneous attention. High-frequency monitoring, enabled by advanced Earth observation constellations, multi-sensor data fusion, and cloud-based analytics, captures wetland dynamics that may vary across daily, seasonal, or interannual timescales. At the same time, mapping must span large spatial extents to generate an integrated understanding of wetland ecosystems at continental or global scales. This dual focus on temporally frequent and spatially extensive observations underpins the detection of subtle shifts in wetland hydrology, vegetation succession, and anthropogenic impacts, ultimately guiding more informed, scalable conservation and restoration strategies.

Truly global, high-frequency wetland monitoring hinges on robust international collaboration. Coordinated efforts to collect and harmonize wetland reference samples, adopt standardized monitoring protocols, and refine multi-source data integration will be crucial in developing consistent, reliable, and timely global wetland products. By leveraging these technological and collaborative frameworks, the remote sensing community can advance toward a future where wetland mapping is not only spatially comprehensive but also temporally agile, thereby underpinning effective global wetland conservation and sustainable management.

\section{Conclusion}\label{sec:conclu}
In this paper, we make a thorough review on the development of remote sensing of wetland classification and mapping. We offer our answers to the questions that arise in Sec. \ref{sec:intro} based on the thorough review and in-depth discussion, respectively. 

\begin{itemize}
    \item What is the scientific importance of wetland classification and mapping? 
    
   \textit{Wetland classification and mapping are instrumental in supporting kinds of scientific research. We list four key scientific supports in Sec. \ref{sec:support}}.
    \item What kinds of data and methods have been used in wetland classification and mapping, especially based on the intersections of remote sensing big data era and AI-driven era?

    \textit{We make a review on the key data and methods in wetland mapping and classification in Sec. \ref{sec:data} and Sec. \ref{sec:method}, including three level-1 data categories and three level-1 method categories.}
    \item What critical insights into wetland changes and driving factors can be derived from wetland classification and mapping research?

    \textit{We summarize four driving factors of wetland changes from wetland classification and mapping research in Sec. \ref{sec:factor}}.
    \item What is the current research paradigm and the limitations?

    \textit{We explain the current research paradigm in Sec. \ref{sec:current}, and then point out limitations in Sec. \ref{sec:gap} and Sec. \ref{sec:limitation} from both method-data aspect and wetland mapping product aspect.}
    \item In face of global climate change and technological innovation, what are the challenges and future opportunities for wetland classification and mapping?

    \textit{We propose the prospects of wetland classification and mapping in Sec. \ref{sec:pros}, including the challenges and the future opportunities.}
\end{itemize}

With the efforts since the 1970s from all over the world, researchers have expanded our knowledge boundaries in this field through continuous advancement in different data acquisition and processing methods. Yet this field still has several consistent challenges remaining ahead:
\begin{itemize}
    \item How can we develop feasible wetland mapping products to meet the needs of different scientific applications, regarding spatial resolutions, temporal frequency and scales? It is crucial to establish a collaborative pathway that co-designs wetland mapping methods and classification systems to effectively address these varied scientific needs.
    \item How can the plenty of multi-source data and emerging computer/remote sensing technologies capture the inherent dynamics and complexity of wetlands more profoundly?
    \item In the context of limited wetland training samples, how can we achieve unsupervised, weakly-supervised, zero-shot, few-shot learning for wetland classification and mapping? 
    \item How can we identify the intrinsic driving factors behind wetland changes occurring across multiple spatial and temporal scales?
\end{itemize}
Now in the era of big data and artificial intelligence, new opportunities arise to further refine existing techniques and address long-standing challenges, paving the way for more accurate and scalable wetland monitoring solutions. Back to the consistent challenges, we believe this paper will help readers to build a comprehensive understanding of wetland classification and mapping, and to find their own answers in this challenging but important research field.

\bibliographystyle{IEEEtran}
\bibliography{refs}

\end{document}